\documentclass{article} % For LaTeX2e
\usepackage[preprint]{template/colm2026_conference}

\usepackage{lineno}

\usepackage{times}
\usepackage{latexsym}

\usepackage[utf8]{inputenc} % allow utf-8 input
\usepackage[T1]{fontenc}    % use 8-bit T1 fonts
\usepackage{hyperref}       % hyperlinks
\usepackage{url}            % simple URL typesetting
\usepackage{booktabs}       % professional-quality tables
\usepackage{amsfonts}       % blackboard math symbols
\usepackage{nicefrac}       % compact symbols for 1/2, etc.
\usepackage{microtype}      % microtypography
\usepackage[table]{xcolor}         % colors
\usepackage{graphicx}       % graphics
\usepackage{multirow}       % multirow tables
\usepackage{array}          % advanced table formatting
\usepackage{rotating}       % rotated tables
\usepackage{adjustbox}      % adjustable boxes
\usepackage{amsmath}
\usepackage{subcaption}

\usepackage{inconsolata}

\usepackage{longtable}
\usepackage{pdflscape}

\definecolor{darkblue}{rgb}{0, 0, 0.5}
\hypersetup{colorlinks=true, citecolor=darkblue, linkcolor=darkblue, urlcolor=darkblue}

\usepackage[most]{tcolorbox}

\newtcolorbox{aiprompt}[1]{
    colback=white,               % Background of the main body
    colframe=black!70,           % Border color (dark gray/black)
    colbacktitle=gray!15,        % Background of the title bar (very light gray)
    coltitle=black,              % Color of the title text
    fonttitle=\bfseries\small,   % Bold and slightly smaller title
    title=#1,                    % Title text
    arc=1.5mm,                   % Subtle rounded corners
    boxrule=0.5pt,               % Thin, professional border
    left=10pt,
    right=10pt,
    top=5pt,
    bottom=5pt,
    fontupper=\ttfamily\small,   % Monospace font for the content
    enhanced,                    % Allows for advanced styling
    attach boxed title to top left={yshift=-2mm, xshift=3mm}, % "Floating" title look
    boxed title style={colframe=black!70, boxrule=0.5pt, arc=1mm}
}

\usepackage{enumitem} % compact lists
\setlist{nosep,leftmargin=*}
\usepackage{caption}  % smaller/tighter captions
\usepackage[normalem]{ulem}  % for \sout strikethrough in change markup

\title{Measuring Competency, Not Performance: Item-Aware\\ Evaluation Across Medical Benchmarks}

\author{
  Zhimeng Luo\textsuperscript{*} \quad
  Lixin Wu\textsuperscript{$\dagger$} \quad
  Adam Frisch\textsuperscript{$\ddagger$} \quad
  Daqing He\textsuperscript{*} \\
  \textsuperscript{*}School of Computing and Information, University of Pittsburgh \\
  \textsuperscript{$\dagger$}Department of Educational Psychology, University of Illinois Urbana–Champaign \\
  \textsuperscript{$\ddagger$}Department of Emergency Medicine, University of Pittsburgh \\
  \texttt{\{zhl123,dah44\}@pitt.edu} \quad
  \texttt{lixinwu2@illinois.edu} \quad
  \texttt{frischan@upmc.edu}
}

\begin{document}
\maketitle

% topic scores reflect coherent, interpretable constructs.

\begin{abstract}

Accuracy-based evaluation of Large Language Models (LLMs) measures benchmark-specific performance rather than underlying medical competency: it treats all questions as equally informative, conflates model ability with item characteristics, and thereby produces rankings that vary with benchmark choice.
To address this, we introduce \textsc{MedIRT}, a psychometric evaluation framework grounded in Item Response Theory (IRT) that (1) jointly models latent competency and item-level difficulty and discrimination, and (2) includes benchmark integrity validation to ensure items within each topic measure a single, coherent underlying ability.
We prospectively evaluate 71 diverse LLMs on a USMLE-aligned benchmark across 11 medical topics. 
As internal validation, \textsc{MedIRT} correctly predicts held-out LLM responses on unseen questions with 83.3\% accuracy. As external validation, IRT-based rankings outperform accuracy-based rankings across 6 independent external medical benchmarks—including expert preferences, holistic clinical tasks, safety judgments, and open-ended queries—achieving 4 wins, 0 losses, and 18\% lower variance. 
As a substantive finding, topic-level competency profiles expose striking domain-specific heterogeneity that aggregate accuracy masks.
As a diagnostic tool, difficulty-tier analysis reveals two distinct response profiles (difficulty-sensitive responding and difficulty-insensitive responding) that require fundamentally different interventions.
These results establish item-aware psychometric evaluation as a more valid and stable foundation for assessing LLMs in medicine, with potential implications for any high-stakes domain where benchmark integrity can be validated, and items vary meaningfully in difficulty and discrimination.

\end{abstract}

\section{Introduction}

Large Language Models (LLMs) have demonstrated remarkable capabilities in high-stakes domains like medicine, achieving expert-level performance on standardized medical licensing exams~\citep{singhal2023, Nori2023}. This progress has been tracked by a growing ecosystem of benchmarks for different levels of expertise, such as MedQA~\citep{Jin2021} and MedMCQA~\citep{Pal2022}, and MedXpertQA~\citep{zuo2025medxpertqa}. However, while existing benchmarks~\citep{shah2025holistic} are valuable, the prevailing paradigm for evaluating these models is built on aggregate accuracy, a metric that masks critical nuances and faces three fundamental challenges that threaten its long-term viability.

First, accuracy assumes all questions are equally informative, yet items inherently vary in difficulty and diagnostic power: a model that answers 20 hard questions correctly is not equivalent to one that answers 20 easy ones, but accuracy cannot distinguish between them. Second, without a unified measurement scale, scores are incomparable across benchmarks: a 70\% on MedQA and a 70\% on MedXpertQA do not reflect the same level of capability. Third, accuracy provides no mechanism to verify \textit{construct validity}~\citep{alaa2025medical}---whether questions grouped under the same topic actually measure a single, coherent underlying ability. Without this check, topic scores may conflate distinct skills, and no amount of additional benchmarks can substitute for it.

% First, the reliance on overall accuracy assumes all questions are equally informative, yet test items inherently vary in quality, difficulty, and diagnostic power. This approach fails to distinguish between models that master difficult concepts and those that merely solve easier problems. Second, this item-level oversight creates a system-level crisis: without a unified scale, scores are incomparable across different benchmarks. Third, without validation of what underlying capabilities are being measured (what psychometricians call ``constructs''—the latent abilities like reasoning, knowledge retrieval, or generalization that we intend to assess), score interpretation becomes arbitrary. This forces the field into an untenable position where model evaluation requires exhaustive reporting across dozens of benchmarks, each measuring potentially overlapping or undefined capabilities.

% To make this concrete: consider two LLMs that both answer 60 out of 100 medical questions correctly---identical accuracy. Model~A, however, answers 20 hard questions correctly and misses most of the easy ones. Model~B answers 20 easy questions correctly and misses nearly all the hard ones. A simple accuracy score treats these profiles as equivalent, yet any clinician would regard them as fundamentally different in capability. An evaluation framework that cannot distinguish them is insufficient for high-stakes medical deployment.

A principled evaluation framework should therefore do more than summarize performance on a fixed benchmark. It should capture stable, generalizable patterns of model competency---supporting reliable expectations about how a new LLM will behave on new questions from the same domain, and producing rankings that remain consistent across diverse evaluation contexts. Critically, it should also provide diagnostic depth: distinguishing whether a model's weaknesses reflect genuine knowledge gaps or superficial sensitivity to question format. These requirements motivate a fundamental shift from performance measurement to competency measurement: from asking
\textit{how many questions did this model get right?} to asking
\textit{what does this model actually know, and where does it fail?}

\begin{figure*}[!htbp]
    \centering
    \includegraphics[width=\textwidth]{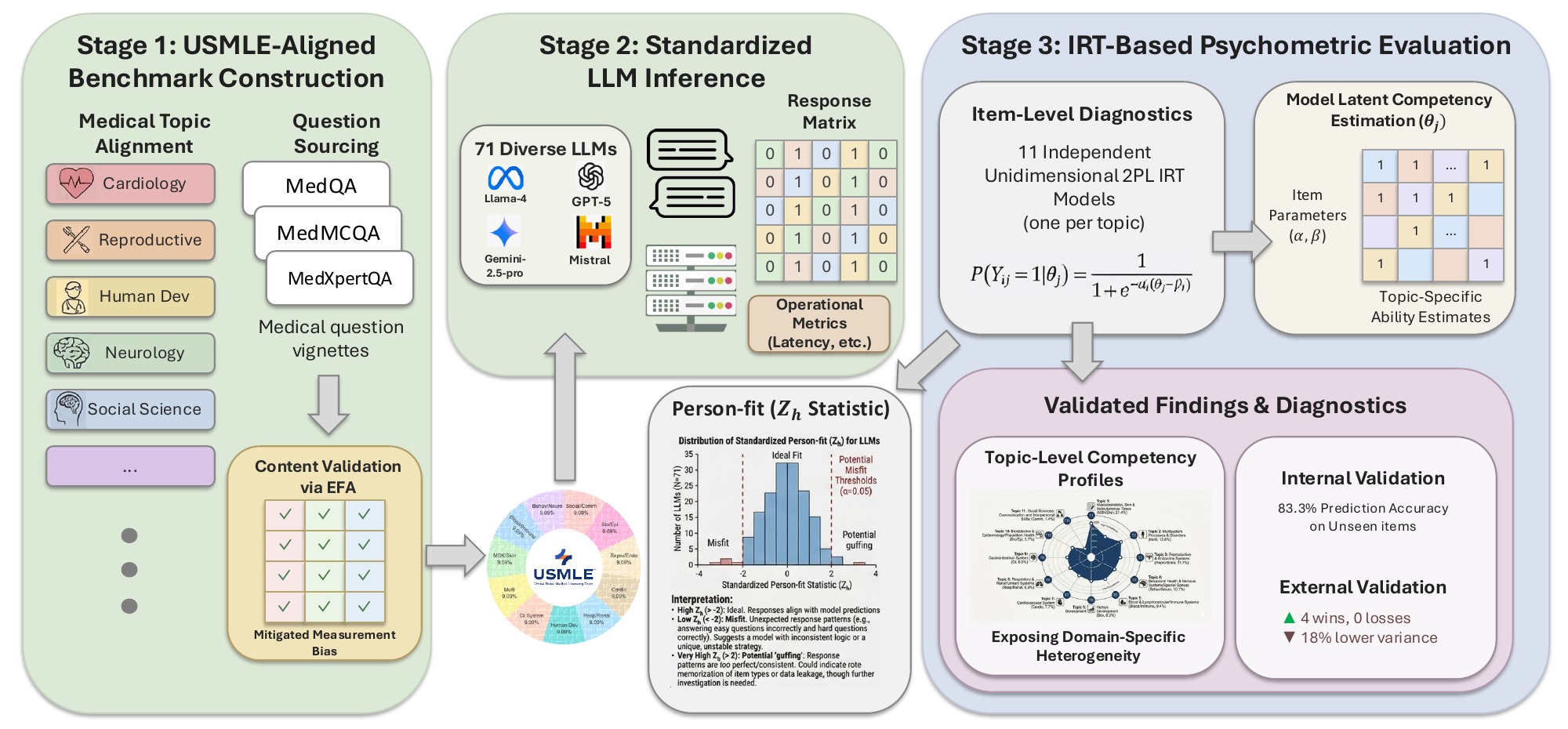}
    \caption{\textbf{An Overview of \textsc{MedIRT} Framework}, illustrating the three integrated components: (1) a USMLE-aligned benchmark with EFA-based content validation, (2) a large-scale LLM cohort, and (3) topic-level 2PL IRT modeling.}
    \label{fig:medtops_framework}
\end{figure*}

To overcome these blind spots, recent work has increasingly turned to Item Response Theory (IRT), the gold standard for high-stakes educational and psychological testing \citep{lord1968statistical}. Unlike simple accuracy, IRT provides a statistical framework to model the interaction between a subject's latent ability and the properties of individual test items (i.e., questions), co-estimating model capability and item difficulty on a unified, continuous scale. However, existing applications of IRT in machine learning often diverge from its core purpose or contain critical methodological flaws. Early work utilized IRT for dataset curation rather than to measure model capabilities \citep{lalor2016building}, while other influential studies built IRT models on archival, outdated data \citep{rodriguez2021evaluation, polo2024tinybenchmarks}. More recent applications have misapplied psychometric models to improve efficiency, such as wrongly using unidimensional IRT to assess abilities across multiple distinct domains \citep{zhuang2023efficiently, zhou2025lost}. Even sophisticated approaches, such as using Multidimensional IRT (MIRT) for LLM routing \citep{song2025irt}, can suffer from limited sample sizes and models with weakly interpretable, data-driven latent dimensions. This leaves a critical gap for healthcare generative AI: the lack of prospective, domain-specific evaluation built from the ground up to provide deep, diagnostic measurement of clinical knowledge in LLMs.

\textbf{Why IRT, specifically?}
Simpler remedies fall short in different ways: multi-benchmark ensembling partially stabilizes rankings but cannot distinguish knowledge gaps from format sensitivity, nor verify that averaged topics measure coherent abilities; Elo-style ratings~\citep{chiang2024chatbot,wu2026medarena} offer preference-calibrated rankings but require dense pairwise comparisons that do not generalize to new models without full re-evaluation.
IRT is the only framework that simultaneously (1)~jointly estimates model ability and item difficulty on a shared scale; (2)~provides a generative model of model-item interactions supporting prediction on unseen items; and (3)~enables construct validation via exploratory factor analysis (EFA), ensuring each topic score reflects a coherent latent ability rather than a heterogeneous mixture of skills.

We introduce \textsc{MedIRT}, a framework that applies IRT prospectively to fresh LLM response data, ensuring findings reflect current model capabilities rather than archival snapshots. To ensure topic scores correspond to coherent latent constructs, we integrate an EFA-based integrity screen that validates topic-wise unidimensionality and prunes weakly aligned items before fitting any IRT model--- thereby directly addressing a core failure mode in medical benchmarking: conflating heterogeneous skills into a single score without verifying that the underlying construct is measurable. We construct an EFA-validated benchmark aligned with \textit{USMLE content specifications}~\citep{USMLE_Content_Outline} and evaluate 71 LLMs, collecting fresh response data via a standardized API protocol through OpenRouter~\citep{OpenRouter2025}. The resulting framework yields topic-specific ability profiles across 11 medical topics, providing a diagnostic foundation for model evaluation in medical education, licensing, and clinical deployment.

Our work makes four principal contributions:

\begin{itemize}
\item We introduce \textsc{MedIRT}, a topic-wise evaluation methodology grounded in IRT that integrates EFA-based benchmark integrity validation, providing more stable and interpretable rankings than accuracy-based measures.

\item We collect prospective response data from 71 diverse LLMs via a standardized API protocol on an EFA-validated benchmark, avoiding the archival data limitations that affect prior work.

\item We validate \textsc{MedIRT} on two independent criteria: held-out prediction accuracy (83.3\% on unseen model-item pairs) and ranking alignment with diverse external benchmarks, demonstrating both internal generalization and external validity.

\item Our analysis reveals two substantive empirical findings: topic-level ability profiles expose striking domain-specific heterogeneity masked by aggregate rankings, and item-level diagnostics distinguish two response modes (difficulty-sensitive and difficulty-insensitive responding) that require fundamentally different interventions. An interactive leaderboard integrating these findings is available at the link in the footnote.\footnote{\url{https://huggingface.co/spaces/anonymized_link}}
\end{itemize}

% \item Our large-scale analysis reveals that medical capability is highly non-uniform across domains. We synthesize our analytical findings into a practical decision support framework, delivered as an interactive platform\footnote{An interactive leaderboard with the benchmark is available at \url{https://huggingface.co/spaces/anonymized_link}.}, that integrates our competency profiles with operational metrics. This provides practitioners with a clear, evidence-based pathway to ensure safe, effective, and trustworthy deployment for specific medical applications.

\section{\textsc{MedIRT} Framework}

\textsc{MedIRT} is a psychometric evaluation framework built around four integrated components, illustrated in Figure~\ref{fig:medtops_framework}:
(1) a 1,100-item benchmark stratified across 11 USMLE-aligned topics;
(2) a standardized inference protocol applied uniformly to all evaluated LLMs;
(3) an EFA-based integrity screen that validates unidimensionality within each topic before any model is fitted;
and (4) 11 independent 2PL IRT models (one per topic) that jointly estimate item difficulty and model ability, producing scores calibrated for question characteristics rather than raw response counts. Together, these components yield topic-specific ability profiles grounded in validated latent constructs, supporting principled model selection for domain-specific medical deployment.

\subsection{The Medical Knowledge Benchmark with USMLE-Aligned Topics}

We constructed a topic-balanced evaluation dataset by integrating 7,906 questions from three established medical benchmarks: MedQA (Test), MedMCQA (Dev), and MedXpertQA (Test). To ensure topic-specific evaluation aligned with medical education standards, questions were classified into 11 medical topics based on \textit{USMLE Step 1 Test Content Specifications}~\citep{USMLE_Content_Outline} using \texttt{GPT-oss-120b}~\citep{agarwal2025gpt}. The classification was verified by one clinician, resulting in a 1\% labeling error rate. Full topic distributions and abbreviations are provided in Appendix Section~\ref{app:dataset_distribution}.

% Initially, we constructed a topic-balanced evaluation dataset by integrating 7,906 questions from three established medical benchmarks: MedQA (Test), MedMCQA (Dev), and MedXpertQA (Test). To ensure topic-specific evaluation aligned with medical education standards, questions were classified according to USMLE Step 1 Test Content Specifications using \texttt{GPT-oss-120b}~\citep{agarwal2025gpt} (given its strong MedQA performance~\citep{vals2025medqa} at lower cost than proprietary models, outperforming \texttt{MedGemma-27b}~\citep{sellergren2025medgemma}). Verified by one domain expert, the topic labels achieve a very low error rate of 1\%, indicating the quality of the automatic labeling for our dataset. Table \ref{tab:dataset_distribution} in Appendix~\ref{sec:} presents the topic distribution across the 11 topics identified in our original dataset prior to balanced sampling.  

% From this labeled pool, we then created a stratified dataset by sampling exactly 100 questions from each of the 11 USMLE-aligned topics, yielding 1,100 total items. This balanced design serves two purposes: (1) it prevents score distortion from topic imbalance: without stratification, a dataset with 500 cardiology and 50 neurology items would primarily measure cardiology knowledge, invalidating claims about ``overall medical capability''; and (2) it ensures sufficient per-topic data (n=100) for stable parameter estimation in both validation analyses and IRT modeling, as these methods require adequate response variation to distinguish item properties from model abilities.

From this labeled pool, we created a stratified dataset by sampling exactly 100 questions from each topic, yielding a total of 1,100 items. This balanced design serves two purposes: (i) it prevents score distortion from topic imbalance
% : without stratification, a dataset with 500 cardiology and 50 neurology items would primarily measure cardiology knowledge, invalidating claims about ``overall medical capability''
and (ii) it ensures sufficient per-topic data ($n=100$) for stable parameter estimation in both EFA and IRT modeling.

\subsection{Validation Procedure via Exploratory Factor Analysis}
\label{sec:efa}

Before fitting IRT models, we validated \textit{unidimensionality}: within each medical topic, items should measure a single coherent capability, as assumed by the 2PL model. If violated (for example, if ``cardiology'' items mix anatomical recall, ECG interpretation, and pharmacology), the resulting topic score becomes uninterpretable, undermining both targeted diagnosis of model weaknesses and meaningful cross-model comparisons.

We applied EFA separately to each topic's item response matrix using \texttt{fa()} from the \texttt{psych} package \citep{Revelle2024} in R~(v4.4.0; \citealt{RCoreTeam2024}), using tetrachoric correlations as input to avoid attenuation of latent associations between binary items. Items were retained if their loading on the single unrotated factor exceeded $0.30$, implying at least 9\% shared variance with the dominant dimension \citep{hair2019multivariate}; negatively loading items were excluded regardless of magnitude.
% Full estimation details, rotation rationale, and edge-case handling are provided in Appendix~\ref{app:efa_details}.

As shown in Table~\ref{tab:dataset_distribution}, most topics retained 55--69 items after screening. The Communication Skills topic retained only 31 items, reflecting the inherently heterogeneous nature of interpersonal competencies; ability estimates for this domain should be interpreted with caution. Across all 11 topics, 645 items passed the unidimensionality screen and were carried forward into IRT calibration. This validation ensures each topic score reflects a coherent latent construct rather than a heterogeneous mixture of skills, providing a principled foundation for the ability estimates reported in Section~\ref{sec:results}.

\subsection{Large Language Model Cohort}

We randomly sampled 100 LLMs from the OpenRouter API \citep{OpenRouter2025} to represent the current generative AI landscape, spanning model sizes from 3B to over 400B parameters, both proprietary and open-source systems, and a range of specializations including general-purpose, instruction-tuned, reasoning-optimized, and code-focused variants. Detailed model statistics are provided in Appendix Table~\ref{tab:model_statistics}.

To ensure reliable evaluation, we applied strict selection criteria: accessible via a standardized API interface, an error rate below 5\%, inference time under 120 seconds per query, and consistent availability throughout the evaluation period. After applying these criteria, 71 models completed the full evaluation protocol and were included in the final analysis. All models were evaluated under identical inference parameters (temperature~0, maximum 3,000 tokens, low reasoning mode, and up to three retries per query) to ensure that observed differences reflect genuine capability rather than configuration artifacts. Full prompt templates and parsing details are provided in Appendix~\ref{sec:prompt} and~\ref{sec:parsing}.

\subsection{Topic-Level 2PL IRT Modeling}
We treat each of the 11 USMLE topics as separate areas of knowledge and build one IRT model per topic using the two-parameter logistic (2PL) framework. Instead of fitting a single, complex multidimensional model, this design provides stable estimates and clear, topic-level profiles of performance.

Medical competency is inherently multidimensional—clinical training and credentialing evaluate knowledge domain by domain \citep{Downing2003, Holmboe2018}. Our framework mirrors this structure: modeling each of the 11 USMLE-aligned topics independently with unidimensional 2PL models produces ability scores $\theta_{m,t}$ that are directly interpretable for diagnosing per-topic strengths and weaknesses.

This design also reflects a practical constraint. While a single multidimensional IRT model spanning all topics is theoretically appealing, attempts to fit one to our dataset ($N=71$ models, 1,100 items) resulted in estimation instability—a known challenge when dimensions are highly correlated and local item dependencies are present \citep{Reckase1997, Yen1984}. Conversely, forcing a single unidimensional model across all topics risks creating biased composite estimates \citep{Reckase1979}. Topic-level modeling achieves the optimal balance of statistical reliability and clinical interpretability.

\subsubsection{IRT Model Specification}
The 2PL model captures two intuitive properties of each question. First, its \textbf{difficulty} ($b$): a question with a high $b$ value is one that only strong models tend to answer correctly, while a low $b$ question is one that nearly all models get right. Second, its \textbf{discrimination} ($a$): a high-$a$ question sharply separates stronger from weaker models, while a low-$a$ question provides little information about ability differences. A model's \textbf{ability} ($\theta$) is estimated from its full pattern of correct and incorrect answers, weighted by these item properties—so answering hard, highly-discriminating questions correctly contributes more to the ability estimate than answering easy ones.

Formally, for each medical topic $t$, the probability that model $m$ correctly answers item $i$ is:
\begin{gather}
\Pr(X_{imt}=1 \mid \theta_{m,t}) = \sigma\!\big(a_{i,t}(\theta_{m,t}-b_{i,t})\big), \\
\qquad \sigma(x)=\tfrac{1}{1+e^{-x}}.
\end{gather}

Here $\theta_{m,t}$ is the latent ability of model $m$ on topic $t$; $b_{i,t}$ is the item difficulty (the ability level at which a model has a 50\% chance of answering correctly); and $a_{i,t}$ is the discrimination parameter. This framework, widely used in educational and psychological measurement \citep{birnbaum1968some}, provides a principled way to decouple model proficiency from item characteristics. Abilities are estimated on a standardized scale (mean~0, SD~1) via marginal maximum likelihood, ensuring comparability across topics and models.

For an overall summary, we report a composite score by averaging standardized topic-level estimates:
$\hat{\Theta}_{m} = \tfrac{1}{11}\sum_{t=1}^{11} Z_{m,t}$,
where $Z_{m,t}$ is the standardized ability for topic $t$. This unweighted aggregation reflects our balanced benchmark design, assigning equal weight to each medical domain in accordance with USMLE content specifications.

\section{Benchmark Experiments}

\subsection{Experiment Settings}

Our evaluation is grounded in a standardized protocol to ensure that all models are assessed under identical conditions, guaranteeing the reproducibility and fairness of our results. This comprehensive evaluation environment ensures direct comparability across all evaluated models and enables statistically valid performance assessments.

For prompt design, a zero-shot, multiple-choice question (MCQ) format was utilized in the study. The prompt frames the task as a ``closed-book multiple-choice medical exam'' and strictly instructs the model to return \textbf{only one letter} corresponding to the single best option, explicitly forbidding any supplementary words or explanations. The prompt templates used in the study are shown in Section~\ref{sec:prompt} of the Appendix. 
We extracted final answers from the raw text using an automated parsing script that strictly adheres to the specified output format, treating any deviation as an instruction-following failure; details are shown in the Appendix Section~\ref{sec:parsing}.

\subsection{Evaluation Metrics}
\label{sec:metrics}

% Our study compares two fundamentally different approaches to evaluating LLM performance in medical domains. To rigorously assess the quality and validity of these competing evaluation methodologies, we employ a comprehensive validation framework alongside practical deployment metrics. The evaluation structure consists of two distinct levels:

% \textbf{Competing Evaluation Approaches}:

% \textit{Accuracy-based evaluation} provide baseline comparisons with existing benchmarks, calculated as the percentage of correct answers. 

% \textit{IRT-based evaluation} constitutes our proposed alternative evaluation framework. Unlike overall accuracy, IRT jointly models both model ability and item characteristics, producing difficulty-adjusted ability estimates on a standardized scale, with positive values indicating above-average proficiency.

% \textbf{Criteria for Comparing Evaluation Quality}:

The central research question of our study is: Which evaluation approach provides a more valid, reliable, and generalizable assessment of medical LLM capabilities? To answer this question, we employ two independent validation criteria that assess different aspects of evaluation quality:

First, we test whether each method \textbf{generalizes within the benchmark (internal validation)}: \textit{can it predict how a new LLM will respond to unseen questions, given only what it learned from other models and items?} We employ a two-stage holdout design: item characteristics are first estimated using 80\% of the LLM cohort (57 models); each remaining held-out LLM's ability is then estimated on 80\% of the items, and its responses on the held-out 20\% are predicted. Higher predicted accuracy indicates that the method captures stable patterns of model competency rather than overfitting to its calibration sample.

Second, we test whether each method \textbf{generalizes across benchmarks (external validation)}: \textit{do the rankings it produces hold up against entirely independent evaluations the method never saw?} We compare against 13 benchmarks in total: 3 internal benchmarks (MedQA, MedMCQA, MedXpertQA) that constitute \textsc{MedIRT}'s own dataset and are included for comparison only, and 10 external benchmarks spanning substantially different evaluation paradigms—expert and general human preference rankings (LMArena~\citep{chiang2024chatbot}, MedArena~\citep{wu2026medarena}), holistic medical task suites (MedHELM~\citep{bedi2026holistic}), conversational health queries (MEDIC~\citep{kanithi2026mediccomprehensiveevaluationleading} HealthBench~\citep{arora2025healthbenchevaluatinglargelanguage}), safety and ethics evaluations (MEDIC MedSafety), open-ended clinical queries (MEDIC Open-Ended), clinical note generation (ACI Bench~\citep{yim2023aci}), and medical summarization—using Kendall's $\tau_b$ on overlapping model subsets, reported in Table~\ref{tab:stability_comparison}. Benchmark selection details and exclusion criteria are provided in Appendix~\ref{app:crossbench}.

\begin{table*}[!htb]
\centering
\footnotesize
\begin{tabular}{llc ccc}
\toprule
& & &
\multicolumn{3}{c}{\textbf{Kendall's $\tau_b$ $\uparrow$ (p-values)}} \\
\cmidrule(lr){4-6}
\textbf{Benchmark}
& \textbf{Domain (Format)}
& \textbf{$n$}
& \textbf{Accuracy}
& \textbf{IRT(All)}
& \textbf{\textsuperscript{\dag}IRT (pr.)} \\
& & & \textbf{Ranking} & \textbf{Ranking} & \textbf{Ranking} \\
\midrule
\multicolumn{6}{l}{\textbf{\textit{Internal Benchmarks (MedIRT Components)}}} \\
MedQA
& Med Knowledge (MCQ)
& 71
& 0.893 ($<$\!.01)
& \textbf{0.901} ($<$\!.01)
& 0.899 ($<$\!.01) \\
MedMCQA
& Med Knowledge (MCQ)
& 71
& \textbf{0.944} ($<$\!.01)
& 0.928 ($<$\!.01)
& 0.928 ($<$\!.01) \\
MedXpertQA
& Exp Knowledge (MCQ)
& 71
& \textbf{0.761} ($<$\!.01)
& 0.747 ($<$\!.01)
& 0.747 ($<$\!.01) \\
\midrule
\multicolumn{6}{l}{\textbf{\textit{External Benchmarks: Out-of-Domain Validation (Significant, $p < 0.10$)}}} \\
LMArena (Med)
& General Prefer (Pairwise)
& 36
& 0.568 ($<$\!.01)
& 0.590 ($<$\!.01)
& \textbf{0.594} ($<$\!.01) \\
LMArena Expert (Med)
& Exp Prefer (Pairwise)
& 21
& 0.438 ($<$\!.01)
& \textbf{0.476} ($<$\!.01)
& \textbf{0.476} ($<$\!.01) \\
MedHELM (All)
& Holistic Tasks (Mixed)
& 6
& 0.733 (.05)
& 0.733 (.05)
& 0.733 (.05) \\
MEDIC (HealthBench)
& Chat (LLM-Rubric)
& 8
& 0.546 (.06)
& 0.546 (.06)
& 0.546 (.06) \\
MEDIC (MedSafety)
& Safety (LLM-Judge)
& 14
& 0.508 (.01)
& \textbf{0.552} ($<$\!.01)
& \textbf{0.552} ($<$\!.01) \\
MEDIC (Open-Ended)
& Open-Ended (Pairwise)
& 12
& 0.455 (.04)
& \textbf{0.485} (.03)
& \textbf{0.485} (.03) \\
\midrule
\multicolumn{6}{l}{\textbf{\textit{External Benchmarks: Out-of-Domain Validation (Non-Significant, $p > 0.10$)}}} \\
MedArena
& Exp Prefer (Pairwise)
& 3
& 1.000 (.33)
& 1.000 (.33)
& 1.000 (.33) \\
MedHELM (Admin)
& Workflow (Mixed)
& 6
& 0.600 (.13)
& 0.600 (.13)
& 0.600 (.13) \\
MEDIC (ACI Bench)
& Generation (LLM-Judge)
& 14
& \textbf{0.341} (.10)
& 0.319 (.13)
& 0.319 (.13) \\
MEDIC (Note Summ)
& Summary (Cross-Exam)
& 14
& -0.297 (.16)
& \textbf{-0.275} (.19)
& \textbf{-0.275} (.19) \\
\bottomrule
\end{tabular}
\caption{\textbf{Cross-Benchmark Generalization: IRT vs Accuracy-Based Rankings.}
Kendall's $\tau_b$ quantifies ranking correlation between evaluation methods.
Higher values indicate the method's ranking aligns more closely with the benchmark.
All correlations computed on overlapping model subsets ($n$ = number of overlapping models). \textbf{Bold} = best value in that row.
\textsuperscript{\dag}IRT (pr.) = IRT (pruned)}
\label{tab:stability_comparison}
\end{table*}

\section{Results}
\label{sec:results}

% \subsection{Comparing Quality of Evaluation Methods}
\subsection{IRT Rankings Generalize Consistently Than Accuracy Internally and Externally}

% We compare IRT-based and accuracy-based evaluation using two complementary criteria: out-of-sample generalization and alignment with human preference judgments.

\begin{figure}[!htb]
\centering
%--- LEFT minipage: Table 1 ---
\begin{minipage}[b]{0.4\linewidth}
  \centering
  \small
  \begin{tabular}{lcc}
  \toprule
  \textbf{Evaluation} & \multicolumn{2}{c}{\textbf{Pred. Acc.\ $\uparrow$}} \\
  \cmidrule(lr){2-3}
  \textbf{Methods} & \textbf{Macro} & \textbf{Micro} \\
  \midrule
  Accuracy & N/A\textsuperscript{\dag} & N/A\textsuperscript{\dag} \\
  \midrule
  IRT (all)             & 0.812 & 0.812 \\
  \textbf{IRT (pruned)} & \textbf{0.833} & \textbf{0.833} \\
  \bottomrule
  \end{tabular}
  \captionof{table}{\textbf{Internal Prediction Generalization Validation.} Predictive accuracy on held-out LLM-item pairs under an 80/20 train-test split. \textsuperscript{\dag}Accuracy-based evaluation is descriptive and does not define a generative model for predicting model-item interactions.}
  \label{tab:validation_comparison}
\end{minipage}%
\hfill
%--- RIGHT minipage: Figure 2 ---
\begin{minipage}[b]{0.56\linewidth}
  \centering
  \includegraphics[width=0.7\linewidth]{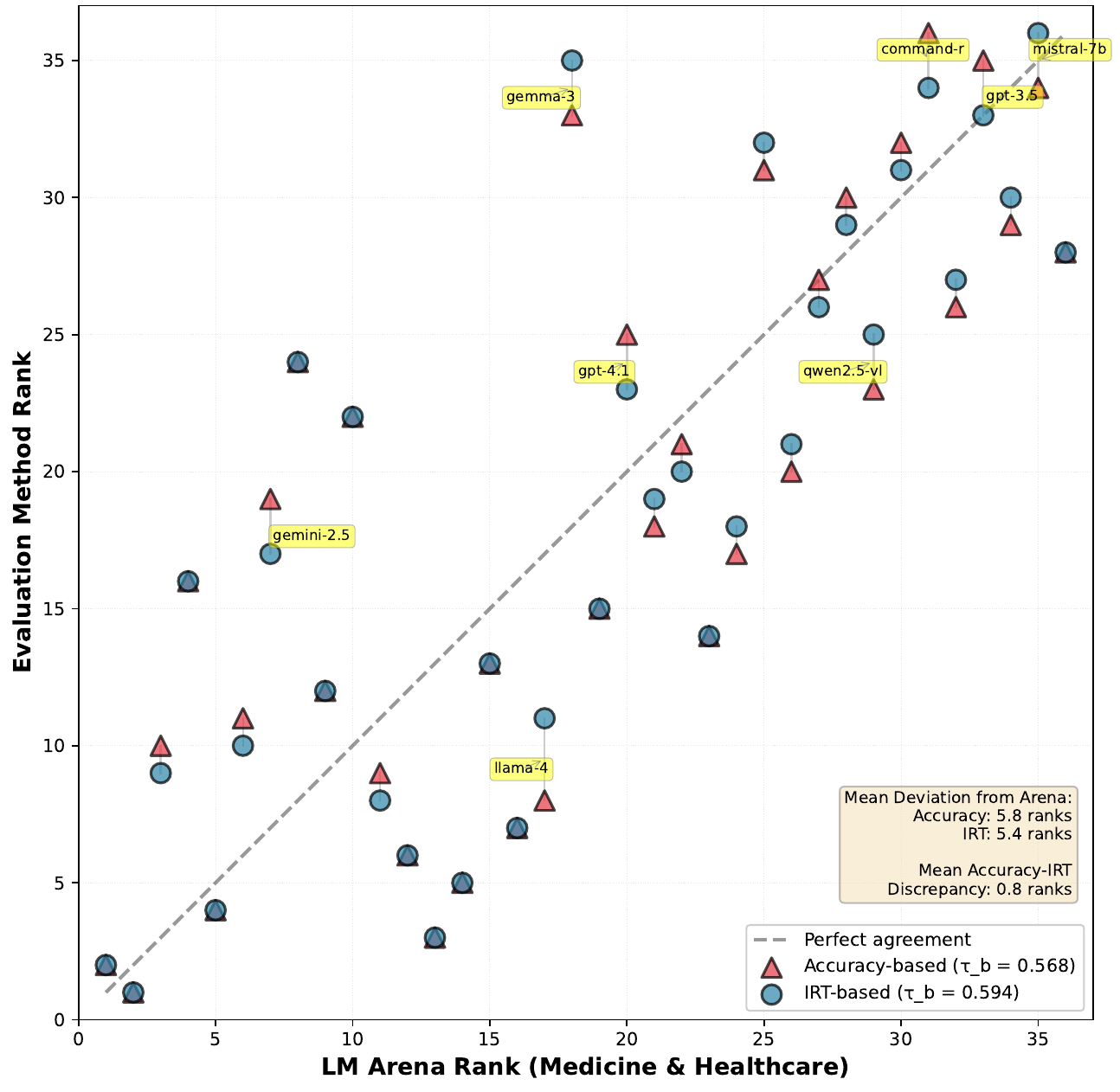}
  \captionof{figure}{\textbf{Ranking Consistency with Human Preference (LM Arena).} Comparison of model rankings from accuracy-based (red triangles) and IRT-based (blue circles) evaluation methods against LMArena rankings (Medicine \& Healthcare).}
  \label{fig:ranking_consistency}
\end{minipage}
\end{figure}

\textbf{Internal validation.}
% A useful evaluation should do more than summarize performance on a fixed benchmark; it should support reliable expectations about how a \emph{new} LLM will behave on \emph{new} questions from the same domain. We test this by asking: after learning from a subset of LLMs and questions, can the method predict held-out responses?
% Table~\ref{tab:validation_comparison} reports predictive accuracy under a two-stage holdout design. We first estimate item characteristics (difficulty and discrimination) using 57 randomly selected LLMs (80\% of the cohort). Holding these item parameters fixed, we estimate each remaining LLM's ability using 80\% of the pruned items and predict outcomes on the held-out 20\%. Under this protocol, IRT achieves a predictive accuracy of 0.833, correctly predicting 83.3\% of held-out LLM--question pairs.
% Notably, the IRT model fitted to the EFA-pruned item set outperforms the model fitted to the original 1{,}100-item benchmark (0.833 vs.\ 0.812). This suggests that psychometric screening removes noisier questions and yields a more stable measurement scale. Overall, these results indicate that IRT captures patterns that generalize beyond the calibration subset.
Table~\ref{tab:validation_comparison} reports predictive accuracy under the two-stage holdout protocol described in Section~\ref{sec:metrics}. IRT achieves a predictive accuracy of 0.833, correctly predicting 83.3\% of held-out LLM-question pairs. Notably, the IRT model fitted to the EFA-pruned item set outperforms the model fitted to the original 1,100-item benchmark (0.833 vs. 0.812), suggesting that psychometric screening removes noisier questions and yields a more stable measurement scale. Macro and micro averaging yield identical values, confirming robustness to class imbalance. Ensemble accuracy partially addresses the ranking instability that motivates this work, but it inherits the same structural limitation as single-benchmark accuracy: it offers no generative model of item-model interactions and cannot predict performance on unseen items. Overall, these results indicate that IRT captures patterns of model competency that generalize beyond the calibration subset.

\begin{figure*}[!htb]
    \centering
    \includegraphics[width=0.9\textwidth]{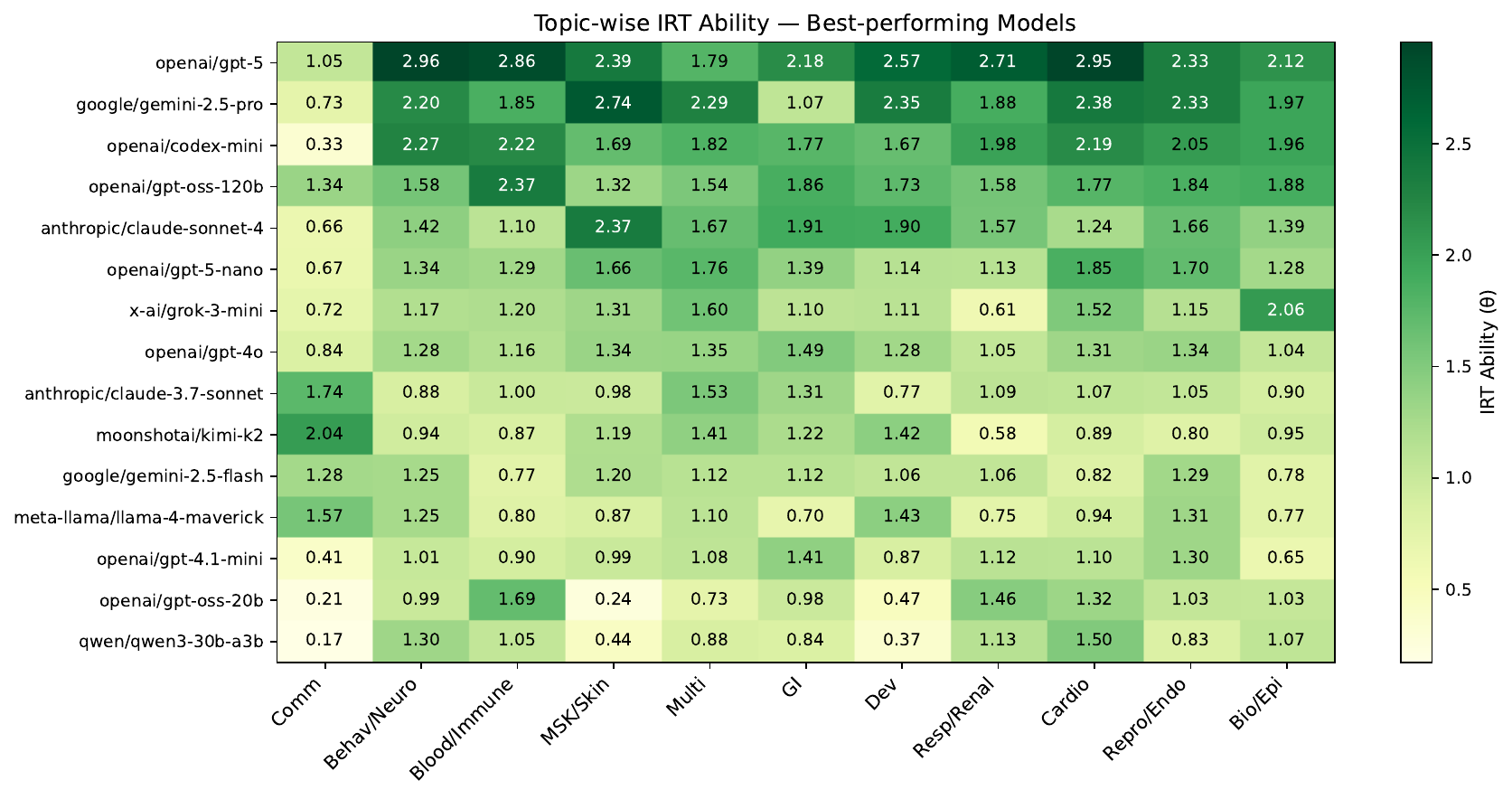}
    \caption{\textbf{Heatmap of topic-wise IRT Ability ($\theta$) derived from pruned item set for the top 15 Models.} Rows list models with the \emph{mean} ability across topics; columns are topic abbreviations, according to Table~\ref{tab:dataset_distribution}.}
    \label{fig:heatmap}
\end{figure*}

\textbf{External validation.}
IRT-based rankings generalize more consistently than accuracy across 10 external benchmarks (Table~\ref{tab:stability_comparison}), none of which were used during \textsc{MedIRT} calibration, spanning expert preference, holistic clinical tasks, safety judgments, and open-ended queries.
On the 6 benchmarks with statistically significant overlap ($p < 0.10$), IRT~(pruned) achieves 4 wins, 2 ties, and 0 losses over accuracy (median $\Delta\tau_b = {+0.026}$), with 18\% lower cross-benchmark variance (SD: $0.087$ vs.\ $0.106$). Overlap sizes of $n = 6$--$36$ models per benchmark preclude individual significance under paired bootstrap tests ($B = 10{,}000$; all $p > 0.10$), a structural constraint of the current evaluation landscape, in which each external leaderboard evaluates a different, largely non-overlapping subset of models.
Directional consistency across all 6 benchmarks (4 wins, 0 losses) is therefore the appropriate evidential standard, and that consistency holds precisely under this structural handicap.

This pattern reflects a systematic difference in what each method captures. Accuracy yields higher internal correlation (mean $\tau_b = 0.866$ vs.\ $0.858$), a predictable consequence of being computed directly from the same items; on external benchmarks the gap reverses (mean $\tau_b = 0.541$ vs.\ $0.564$), indicating that accuracy rankings are partially artifacts of \textsc{MedIRT}'s specific item pool rather than the underlying medical competency construct, a form of in-sample overfitting that IRT's latent-variable formulation avoids.

Construct specificity further validates what \textsc{MedIRT} measures. Correlations are strongest where clinical reasoning is required (expert preference: $\tau_b = 0.476$; safety judgment: $\tau_b = 0.552$; open-ended clinical queries: $\tau_b = 0.485$) and near zero or negative for tasks orthogonal to medical knowledge, such as note summarization ($\tau_b = -0.275$), confirming that the framework captures medical reasoning rather than generic language ability.
That IRT maintains non-inferior rankings across benchmarks that are themselves only weakly correlated (mean $\tau_b = 0.49$, vs.\ $0.77$ among internal components) confirms generalization across genuinely distinct competencies.

\textbf{Model-level rank shifts} help explain this pattern. In Figure~\ref{fig:ranking_consistency}, each point represents one model's rank in the evaluation method (y-axis) versus its rank in LMArena (x-axis). Gray lines show each LLM's rank shift between methods, with yellow lines indicating larger disagreements. Models such as \texttt{llama-4-maverick} and \texttt{gemini-2.5-flash} exhibit notable shifts (highlighted in yellow). For example, \texttt{llama-4-maverick} ranks substantially higher under IRT than under raw accuracy, consistent with strong performance on harder, more discriminating questions that IRT appropriately upweights. The mean absolute deviation from Arena rankings is 5.4 for IRT versus 5.8 for accuracy, further supporting IRT's closer alignment.

Together, these results establish IRT-based rankings as the more stable basis for cross-context model selection in medicine, matching or exceeding accuracy wherever the two diverge and providing diagnostic depth that accuracy structurally cannot.
Cost- and time-normalized ability rankings for the Pareto-efficient model set are reported in Appendix~\ref{sec:cost-performance}.

\subsection{LLMs' Spiky Performance}
\label{sec:topic}

Our topic-level decomposition reveals that no model achieves consistent mastery across all 11 medical specialties. Figure~\ref{fig:heatmap} presents IRT ability estimates ($\theta$) for the top 15 models across all topics, where darker shading indicates higher proficiency. Even among elite performers, pronounced domain-specific variations emerge, challenging the notion of a single universally optimal model.

\textbf{Domain-Specific Leadership and ``Spiky'' Competency Profiles.}
While aggregate rankings suggest a linear hierarchy, our heatmap reveals a complex landscape of distinctive ability ``fingerprints.'' \texttt{GPT-5} exhibits a strong generalist profile, dominating 8 of 11 topics with exceptional scores in Behavioral/Neurological Science ($\theta=2.96$) and Cardiovascular Systems ($\theta=2.95$). However, it is not universally superior. Mid-tier ``specialist'' models frequently outperform the overall leader in specific domains. For instance, \texttt{Gemini-2.5-pro} (ranked 2nd) claims the top spot in Multisystem Processes ($\theta=2.29$) and Musculoskeletal/Skin ($\theta=2.74$). Even more strikingly, \texttt{Kimi-k2} (ranked 12th overall) surpasses \texttt{GPT-5} by nearly one standard deviation in Communication/Interpersonal Skills ($\theta=2.04$ vs. $\theta=1.05$), despite suffering severe deficits in Respiratory/Renal systems ($\theta=0.58$). Similarly, \texttt{Claude-3.7-sonnet} shows pronounced specialization, excelling in Communication ($\theta=1.74$) while underperforming in Developmental Processes ($\theta=0.77$). These ``spiky'' profiles demonstrate that medical competence is fundamentally multidimensional, where overall rankings often mask critical, domain-specific expertise.

\textbf{Systematic weaknesses across the cohort.} The heatmap also exposes domains that pose universal challenges. Communication/Interpersonal Skills shows consistently lower abilities across most models (mean $\theta=0.89$ for top 15), with 11 of 15 models scoring below $\theta=1.0$. Similarly, Developmental Processes (mean $\theta=1.25$) and Biostatistics/Epidemiology (mean $\theta=1.29$) reveal persistent difficulties even after adjusting for item difficulty. These patterns identify critical priorities for future model development and highlight domains where current LLMs may require additional safeguards in clinical deployment.

\subsection{LLMs' Divergent Response Profiles}
\label{sec:failure_modes}

\begin{figure}[!htbp]
  \centering
  \includegraphics[width=\textwidth]{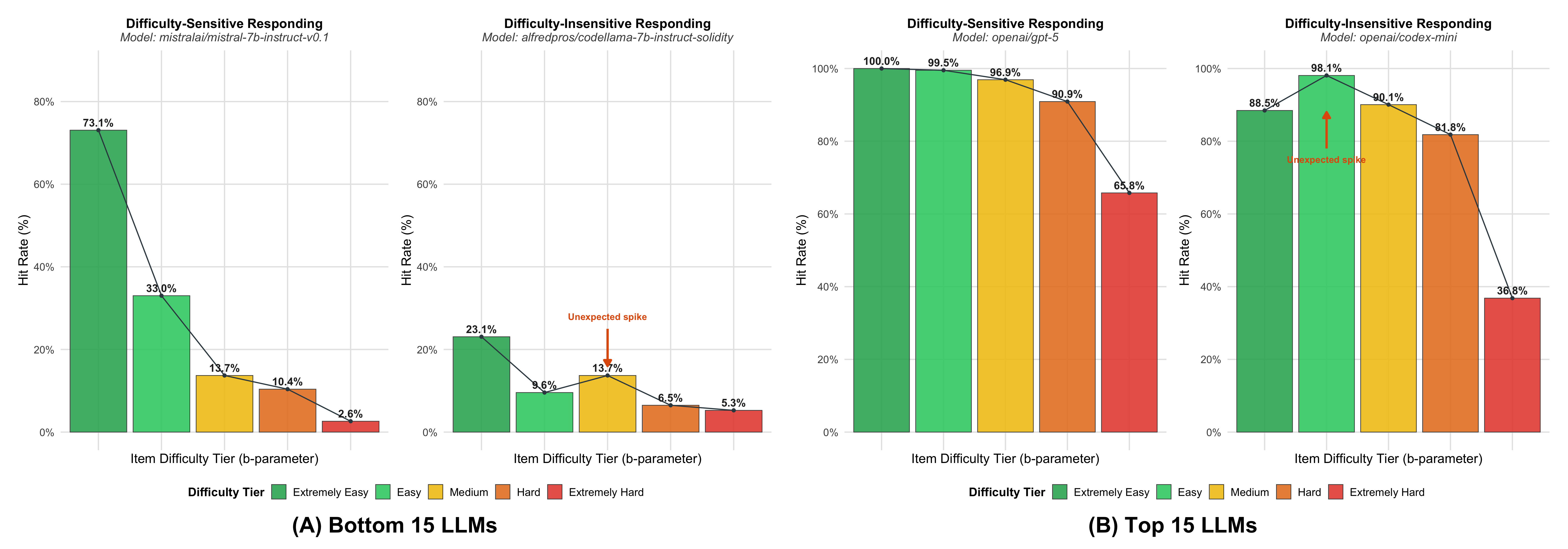}
  \caption{\textbf{Difficulty-Tier Hit-Rate Profiles: Contrasting DSR and DIR Across Ability Strata.}
  Each bar shows the percentage of questions answered correctly within that item difficulty tier, with figure labels (A, B) denoting the bottom-15 and top-15 LLM cohorts respectively. Within each subfigure, the left panel shows a clean monotonic decline. The right panel shows the opposite: hit rates
  rise from one tier to the next before falling again. Note that the two subfigures use different y-axis scales reflecting the distinct ability ranges
  of each cohort.}
  \label{fig:irt_tierprofile_contrast}
\end{figure}

\textbf{Two response profiles emerge across the ability spectrum.}
Competency profiles reveal \textit{where} LLMs struggle; a complete diagnostic account requires also characterizing \textit{how} response accuracy relates to item difficulty. Applying a difficulty-tier inversion diagnostic to the 15 lowest-competency LLMs ($\bar{\theta} < -1$) yields two distinct profiles. The majority exhibit \textbf{difficulty-sensitive responding} (DSR), in which hit rates decline monotonically across difficulty tiers. \texttt{mistral-7b-instruct-v0.1} is a clear example: performance falls steadily from 73.1\% on extremely easy items to just 2.6\% on extremely hard ones, a pattern fully consistent with expected difficulty ordering. A meaningful subset of four LLMs, however, show \textbf{difficulty-insensitive responding} (DIR). \texttt{codellama-7b-instruct-solidity} illustrates this strikingly: despite scoring only 9.6\% on easy items, it recovers to 13.7\% on medium items, passing relatively harder items while failing easier ones, a pattern entirely invisible under raw accuracy. Panel~A of Figure~\ref{fig:irt_tierprofile_contrast} illustrates this contrast.

These two profiles are not confined to low-competency models; they extend across the full range of ability. \texttt{gpt-5} (ranked 1st overall) exhibits DSR, with hit rates declining steadily to 65.8\% on extremely hard items. \texttt{codex-mini}, by contrast, displays a DIR signature: its hit rate \textit{rises} from 88.5\% on extremely easy items to 98.1\% on easy items before declining, an inversion at the ceiling of the leaderboard that aggregate rankings would never surface. Panel~B of Figure~\ref{fig:irt_tierprofile_contrast} illustrates both cases, and a full record of all DIR models is reported in Table~\ref{tab:dir_inversion}. The distinction is not merely descriptive: DSR and DIR models exhibit observably different relationships between item difficulty and response accuracy, and conflating them under a single aggregate score actively misleads deployment decisions.

%%%%%%%% This is the simplifled version of March 31th 2026.
% \textbf{Theoretical grounding.}
% These two profiles, observed across the ability spectrum, are consistent with distinct cognitive mechanisms, though the precise account for any individual model remains an open question. One candidate explanation for DSR is Gagn\'{e}'s \textit{cumulative learning model} \citep{gagne1968contributions, gagne1962factors}: if knowledge is organized as a strict prerequisite hierarchy, failures on foundational items would propagate upward, producing the monotonically increasing error rates with difficulty that DSR exhibits. DIR may reflect a different set of mechanisms. \textit{Fluency misattribution} \citep{whittlesea1993illusions, birch2017curse} offers one account: items closely matching training-time surface patterns may elicit confident but unverified responses, while harder items---lacking a strong pattern match---receive more deliberate processing that happens to succeed. The \textit{curse of knowledge} \citep{camerer1989curse, birch2017curse} provides a related account, whereby a model responds to a familiar surface form rather than to the specific structure of the question as posed. Either mechanism would produce performance contingent on surface format rather than item difficulty---precisely the ordering inversion that the diagnostic detects. These accounts are not mutually exclusive, and disentangling them is beyond the scope of the present analysis; what the diagnostic establishes is the observational profile itself.

\textbf{Theoretical grounding.}
These two profiles are inspired with distinct cognitive mechanisms: DSR aligns with prerequisite-structured knowledge hierarchies \citep{gagne1968contributions, gagne1962factors}, while DIR is consistent with responding driven by surface-format familiarity rather than item difficulty---whether through fluency misattribution \citep{whittlesea1993illusions, birch2017curse} or the curse of knowledge \citep{camerer1989curse, birch2017curse}.
The precise account for any individual model remains an open question; what the diagnostic establishes is the observational pattern, not its underlying cause.
Additional corroboration is provided by the person-fit statistic $Z_h$; distributional evidence and the hard-item hit analysis supporting both DSR and DIR are reported in Appendix~\ref{app:irt_diagnostic}.

\subsection{Practical Implications}

The analyses above translate into three concrete steps for practitioners evaluating medical LLMs. \textit{Running EFA before trusting any topic score is essential}: item retention after screening varied from 31 to 69 items across topics in our benchmark, and scores computed from unvalidated item pools may conflate distinct skills; the improvement in held-out prediction (81.2\%\,$\to$\,83.3\%) confirms that this step meaningfully tightens the measurement scale. \textit{For deployment decisions, topic-wise IRT ability profiles should replace aggregate accuracy}, since a model ranked 12th overall may outperform the top-ranked model by nearly one standard deviation in the specific domain a workflow requires (e.g., \texttt{Kimi-k2} in Communication Skills), a distinction invisible under any aggregate score. \textit{Finally, the inversion diagnostic should be checked before any high-stakes deployment}: a model exhibiting difficulty-insensitive responding (DIR) warrants prompt engineering and adversarial robustness testing regardless of its aggregate estimated $\theta$ rank, because its errors are driven by surface-format sensitivity rather than knowledge depth, a deployment risk that leaderboard rankings structurally cannot surface.

\section{Conclusion}

We introduced \textsc{MedIRT}, a psychometric framework that verifies benchmark integrity via EFA-based item screening and estimates topic-wise abilities using per-topic 2PL IRT models. We apply this framework to 71 diverse LLMs on a USMLE-aligned benchmark covering 11 medical topics.

Our evaluation reveals four key insights. First, Benchmark coherence cannot be assumed, as EFA screening shows substantial variation in topic unidimensionality; the number of retained items ranges from 31 to 69 across topics. Fitting IRT models on this validated item set improves held-out prediction accuracy from 81.2\% to 83.3\%. Second, rankings derived from IRT also outperform those based on raw accuracy, according to external criteria, achieving 4 wins and 0 losses with 18\% lower variance across 6 independent external benchmarks. Furthermore, medical competency is highly non-uniform. LLMs exhibit ``spiky'' profiles that aggregate scores often obscure, where a mid-ranked model can outperform the overall leader by nearly one standard deviation in specific clinical domains. Finally, item-level diagnostics distinguish between two response profiles, difficulty-sensitive responding (DSR) and difficulty-insensitive responding (DIR), which require fundamentally different interventions.

These findings suggest that the information discarded by simple accuracy—item difficulty structure, response-pattern anomalies, and construct coherence—is critical for determining whether a model is suitable for clinical deployment. By recovering this structure, \textsc{MedIRT} transforms evaluation from a ranking exercise into a diagnostic instrument. This approach has implications for any high-stakes domain where benchmark integrity must be validated and items vary in difficulty and discrimination.

\clearpage

% \section*{Limitations}
% This study has several limitations. 
% First, its reliance on static multiple-choice medical questions does not fully capture the dynamic nature of clinical reasoning.

% Second, our evaluation was shaped by practical constraints. Our reliance on the OpenRouter API excluded specialized medical LLMs from our analysis, and the token and time limits we imposed for cost-efficiency may have compromised the full potential of high-capacity reasoning models.
% Third, the domain-specific analysis may not reflect the interdisciplinary integration required in actual medical practice. 

% While LMArena (Medicine \& Healthcare) provides a valuable proxy for human preference, it has inherent constraints as a definitive ground truth. The platform relies on crowd-sourced head-to-head comparisons, where the varying medical expertise of human evaluators may introduce subjective noise into the rankings. Additionally, the Arena rankings reflect a static temporal snapshot that evolves as new models are integrated, potentially affecting longitudinal consistency. Furthermore, there may be a domain mismatch: LMArena likely emphasizes conversational utility and ``bedside manner,'' whereas our benchmark focuses on technical diagnostic accuracy and specialized medical knowledge.

\section*{Ethical Considerations}

This work evaluates the capabilities and limitations of large language models for medical question answering and does not constitute medical advice, clinical decision support, or a deployable diagnostic system.
Accordingly, we emphasize that benchmark performance, including IRT-based ability estimates, should not be interpreted as sufficient evidence for clinical use without domain-scoped validation, safety controls, and appropriate human oversight.

\textbf{Data sources, privacy, and human subjects.}
Our benchmark is constructed from publicly available medical question sources and contains no protected health information (PHI) and no patient-level clinical records.
We do not recruit human participants, and our analyses focus on model outputs generated under controlled inference conditions.

\textbf{Risks of misuse and overgeneralization.}
Model rankings can be misused as marketing claims or as a justification for high-stakes deployment in settings not reflected by the benchmark distribution.
To mitigate this risk, our paper foregrounds diagnostic interpretation (topic-wise profiles and domain weaknesses) and explicitly validates evaluation quality rather than presenting a leaderboard as an end in itself.
We recommend that any real-world use be preceded by local, workflow-specific validation and continuous monitoring, particularly in domains that appear broadly challenging across models (e.g., interpersonal communication or safety-relevant decision-making).

\textbf{Resource use and sustainability.}
Large-scale evaluation of LLMs incurs computational and financial costs. We partially mitigate this impact through standardized, bounded inference settings and by focusing on measurement procedures that maximize diagnostic yield per evaluation query.

\section*{Reproducibility}

All code and processed data used in this study will be released under the MIT License. The release complies with the original licenses of the source datasets.
All datasets used in this work were originally released for research and benchmarking purposes. Our use is limited to offline evaluation of language models and is fully consistent with the intended use specified by the dataset creators. This work does not involve clinical deployment or real-world decision making.

\section*{LLM Usage Disclosure}
This study employs artificial intelligence in two distinct capacities. First, LLMs serve as the primary subjects of our investigation, where we evaluate and rank their performance on medical capabilities. In this capacity, the LLMs are not used as research instruments but rather constitute the object of study itself. Second, we utilized AI-assisted tools (specifically Claude Sonnet 4.6) to refine the manuscript's grammar and polish its language. All substantive intellectual content, including research design, analysis, interpretation, and argumentation, was generated by the human authors. The use of AI for language editing was limited to improving clarity, correcting grammatical errors, and enhancing readability, without altering the scientific content or conclusions of our work.

\bibliography{colm2026_conference}
\bibliographystyle{template/colm2026_conference}

\clearpage
\appendix

\section{Implementation Details}

\subsection{Prompt Templates}
\label{sec:prompt}

We provide here the exact user prompt template used to query models:

\begin{aiprompt}{Standardized Prompt}
You are taking a closed-book multiple-choice medical exam. \\
Answer with ONLY ONE LETTER from [Answer Letters] corresponding to the single best option, and NOTHING ELSE. Do not include any words, punctuation, or explanation. \\
\\
Question: \{question\_text\} \\
\\
Options: A. \{option\_1\} B. \{option\_2\} \ldots \\
\\
Answer:
\end{aiprompt}

\subsection{Answer Extraction from Raw LLM Outputs}
\label{sec:parsing}

To automatically extract multiple-choice answers from diverse LLM response formats, we implement a hierarchical pattern-matching system that categorizes extraction confidence into ``strong'' and ``structured'' evidence levels. Strong patterns capture explicit answer markers such as XML-style tags (\texttt{<answer>B</answer>}), labeled declarations (e.g., ``Answer: C'', ``The correct answer is D''), and markdown-emphasized single letters, while structured patterns identify format-based conventions including leading option labels (``A. Text''), bracketed tokens (``[B]''), and isolated letter tokens on single lines. The system normalizes text through case-folding and Unicode standardization, then applies pattern matching with validation logic that detects conflicts (multiple distinct candidates), ambiguities (structured patterns disagreeing with strong patterns), and failure modes including refusal language, excessive unknown tokens, and safety-marker interference. We prioritize strong evidence when available but require consistency across pattern types, returning an error status for any output exhibiting conflicting signals, multiple answer candidates, or absence of recognizable answer formats. This conservative approach minimizes false positive extractions while accommodating the substantial format diversity observed in contemporary LLM outputs across different prompting strategies and model families.

\begin{table*}[!hb]
\centering
\small
\begin{adjustbox}{width=\textwidth,center}
\begin{tabular}{llrrr}
\toprule
\textbf{Medical Topic} & \textbf{Abbr.} & \textbf{\#Samples (\%)} & \textbf{\#Pruned} \\
\midrule
Musculoskeletal, Skin \& Subcutaneous Tissue & MSK/Skin & 1,693 (21.4\%) & 59 \\
Multisystem Processes \& Disorders & Multi & 1,012 (12.8\%) & 60 \\
Reproductive \& Endocrine Systems & Repro/Endo & 926 (11.7\%) & 61 \\
Behavioral Health \& Nervous Systems/Special Senses & Behav/Neuro & 849 (10.7\%) & 63 \\
Blood \& Lymphoreticular/Immune Systems & Blood/Immune & 743 (9.4\%) & 62 \\
Human Development & Dev & 658 (8.3\%) & 58 \\
Cardiovascular System & Cardio & 606 (7.7\%) & 66 \\
Respiratory \& Renal/Urinary Systems & Resp/Renal & 545 (6.9\%) & 61 \\
Gastrointestinal System & GI & 471 (6.0\%) & 55 \\
Biostatistics \& Epidemiology/Population Health & Bio/Epi & 294 (3.7\%) & 69 \\
Social Sciences: Communication and Interpersonal Skills & Comm & 109 (1.4\%) & 31 \\
\midrule
\textbf{Total} &  &\textbf{7,906} & \textbf{1,100} \\
\bottomrule
\end{tabular}
\end{adjustbox}
\caption{\textbf{Distribution of Medical Topics}, according to the \emph{USMLE Step 1 Test Content Specifications}, labeled by \texttt{GPT-oss-120b} model. Abbr. refers to abbreviations of each topic used in later sections.}
\label{tab:dataset_distribution}
\end{table*}

\section{Benchmark Details}

\subsection{Distribution of Medical Topics}
\label{app:dataset_distribution}

Table~\ref{tab:dataset_distribution} presents the full distribution of questions across the 11 USMLE Step 1-aligned medical topics before balanced sampling, together with the number of items retained per topic after EFA-based pruning (see Section~\ref{sec:efa}). Topic abbreviations used throughout the paper are defined in the \textbf{Abbr.} column. Questions were classified using \texttt{GPT-oss-120b}~\citep{agarwal2025gpt}, selected for its strong MedQA performance~\citep{vals2025medqa} at lower cost than proprietary models, outperforming \texttt{MedGemma-27b}~\citep{sellergren2025medgemma} as the labeling model. Note that item retention after pruning varied substantially across topics---from 31 items (Communication Skills) to 69 items (Biostatistics/Epidemiology)---reflecting meaningful differences in the degree to which each topic measures a single coherent underlying ability.

\subsection{External Benchmark Selection}
\label{app:crossbench}

We screened publicly available medical LLM leaderboards for inclusion in the cross-benchmark generalization study (Table~\ref{tab:stability_comparison}), applying two inclusion criteria: (i) a minimum of $n \geq 3$ overlapping models with MedIRT's 71-model cohort after name harmonization across leaderboard formats, and (ii) publicly accessible rankings from official sources or published papers to ensure reproducibility.

From an initial pool of over a dozen candidate benchmarks, we excluded the Open Medical-LLM Leaderboard (outdated; no overlapping models) and several benchmarks cited in published work but lacking publicly accessible leaderboard rankings. Beyond the three benchmarks embedded in MedIRT (MedQA, MedMCQA, MedXpertQA) and the original LMArena Medical benchmark, we incorporated 9 external benchmarks: two expert preference benchmarks---LMArena Expert Medical (filtered from the full LMArena dataset by occupational tag to retain expert-level raters only) and MedArena (restricted to certified clinicians; leaderboard retrieved February~18, 2026)---two holistic medical task benchmarks (MedHELM comprehensive and administration/workflow subtasks), and five MEDIC benchmarks covering conversational health queries (HealthBench), safety and ethics evaluation (MedSafety), open-ended clinical queries, clinical note generation (ACI Bench), and medical summarization.

The substantially lower inter-correlation among the 6 significant external benchmarks (mean $\tau_b = 0.49$) compared to MedIRT's internal components (mean $\tau_b = 0.77$) confirms that these benchmarks assess genuinely distinct competencies rather than redundant measurements, validating the breadth of the generalization evidence.

\begin{figure}[!htb]
\centering
%--- LEFT minipage: Table 1 ---
\begin{minipage}[b]{0.48\linewidth}
    \centering
    \includegraphics[width=\textwidth]{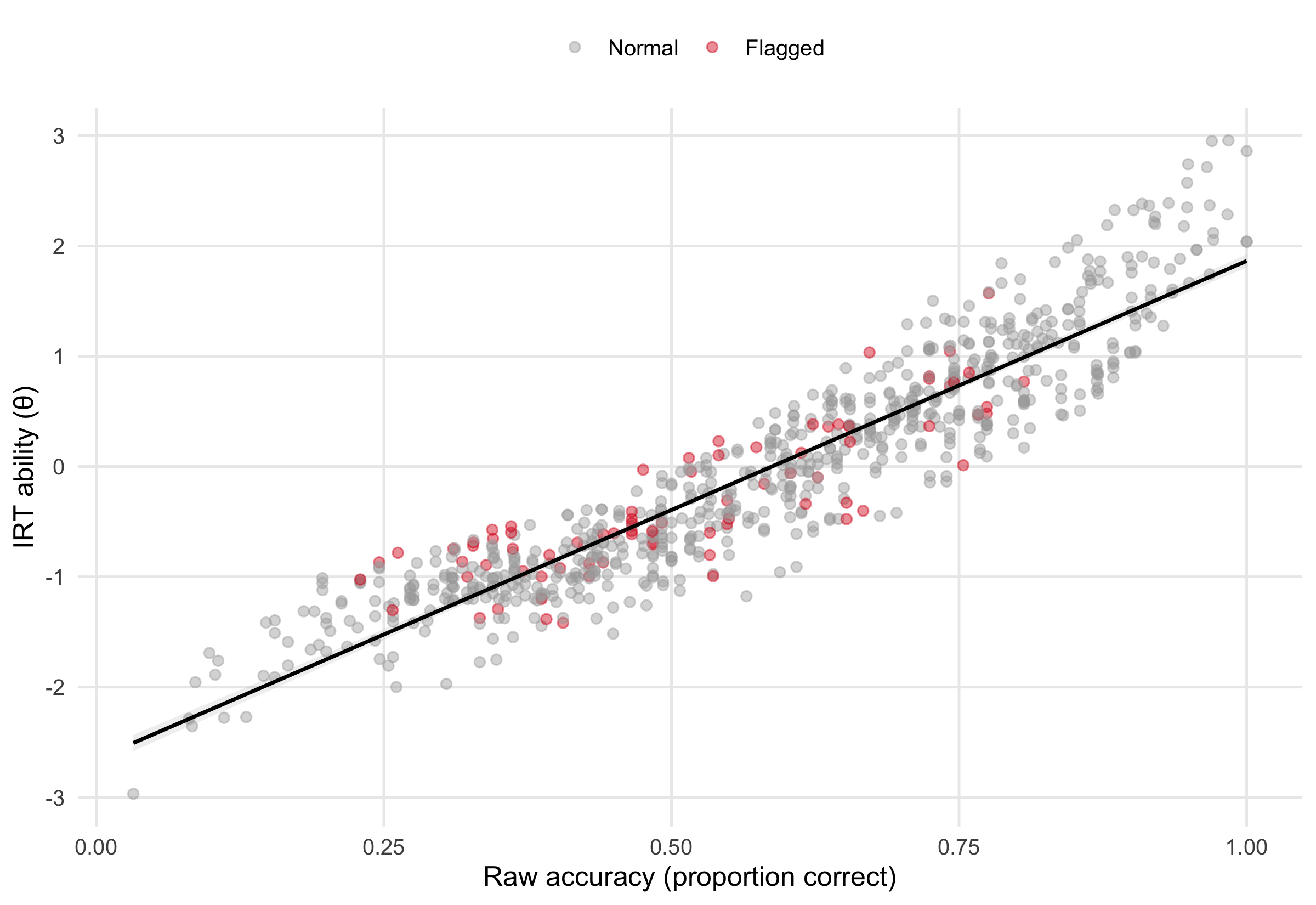}
    \caption{IRT ability ($\theta$) vs.\ raw accuracy per model--topic pair. Wide vertical spread at any given accuracy level indicates that IRT captures information beyond proportion correct. Red points are formally misfitting pairs (Zh $< -1.96$).}
    \label{fig:theta_vs_accuracy}
\end{minipage}%
\hfill
%--- RIGHT minipage: Figure 2 ---
\begin{minipage}[b]{0.48\linewidth}
  \centering
  \includegraphics[width=\textwidth]{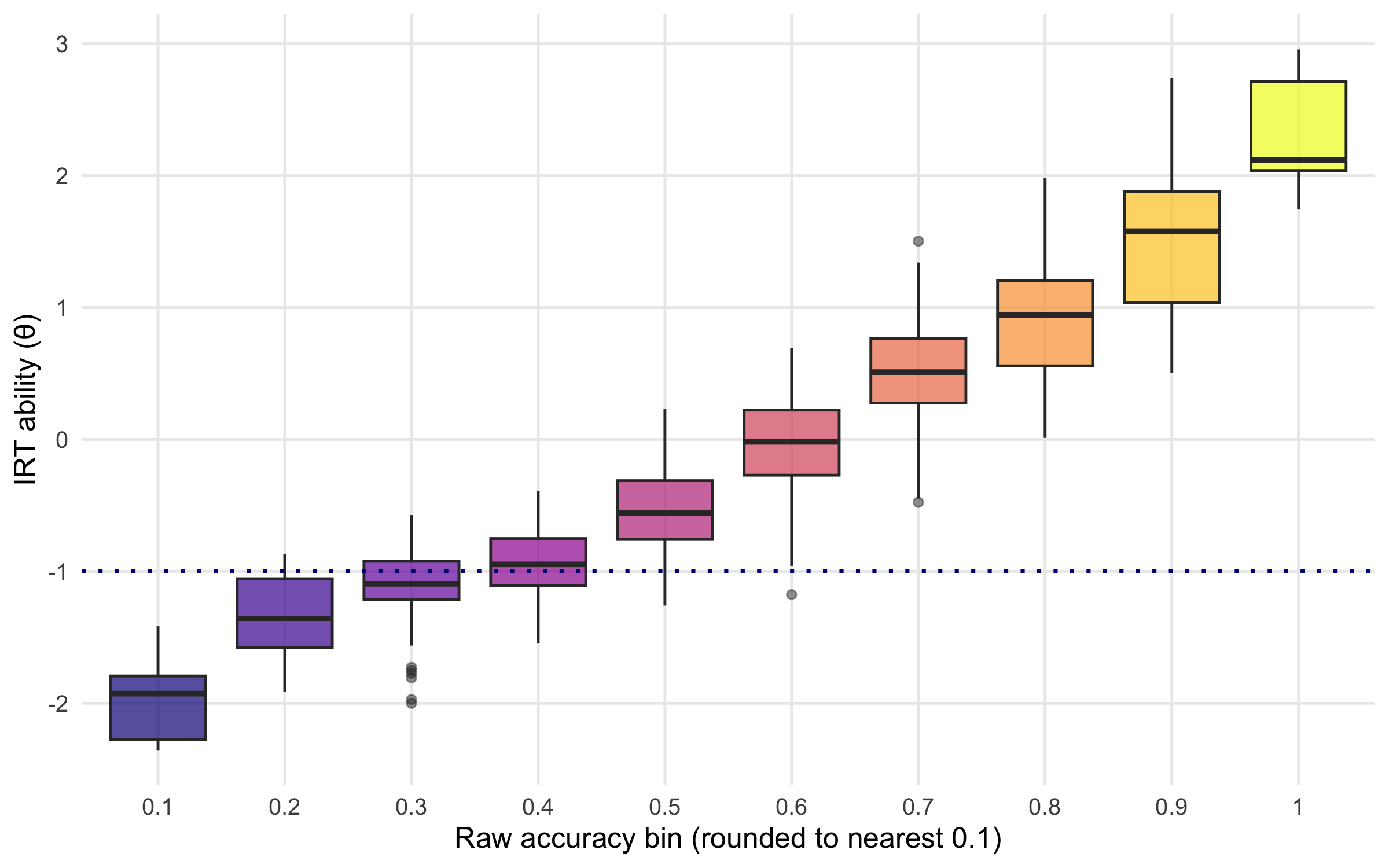}
  \caption{IRT ability ($\theta$) spread within raw accuracy bins. Wide interquartile ranges confirm substantial within-bin $\theta$ variance. The blue dashed line marks $\theta = -1$, the baseline competency threshold.}
  \label{fig:theta_spread_bins}
\end{minipage}
\end{figure}

\section{Extended Diagnostic Evidence}
\label{app:irt_diagnostic}

This appendix provides visual evidence to support the IRT-based analyses in Sections~3 and~4 for readers seeking additional methodological detail.

\paragraph{IRT ability adds signal beyond raw accuracy.}
Figures~\ref{fig:theta_vs_accuracy} and~\ref{fig:theta_spread_bins} together demonstrate that $\theta$ and raw accuracy, while correlated, are not interchangeable. At any fixed accuracy level, $\theta$ estimates span a range of up to two logit units, reflecting differences in \textit{how} models answer — specifically, whether their correct and incorrect responses align with the expected item difficulty ordering. This within-accuracy variation is the information IRT recovers that proportion-correct discards. Flagged model--topic pairs (Zh $< -1.96$, shown in red in Figure~\ref{fig:theta_vs_accuracy}) appear across the full accuracy range, confirming that response-pattern anomalies are not detectable from accuracy alone.

\begin{figure}[!htb]
\centering
%--- LEFT minipage: Table 1 ---
\begin{minipage}[b]{0.48\linewidth}
    \centering
    \includegraphics[width=0.8\textwidth]{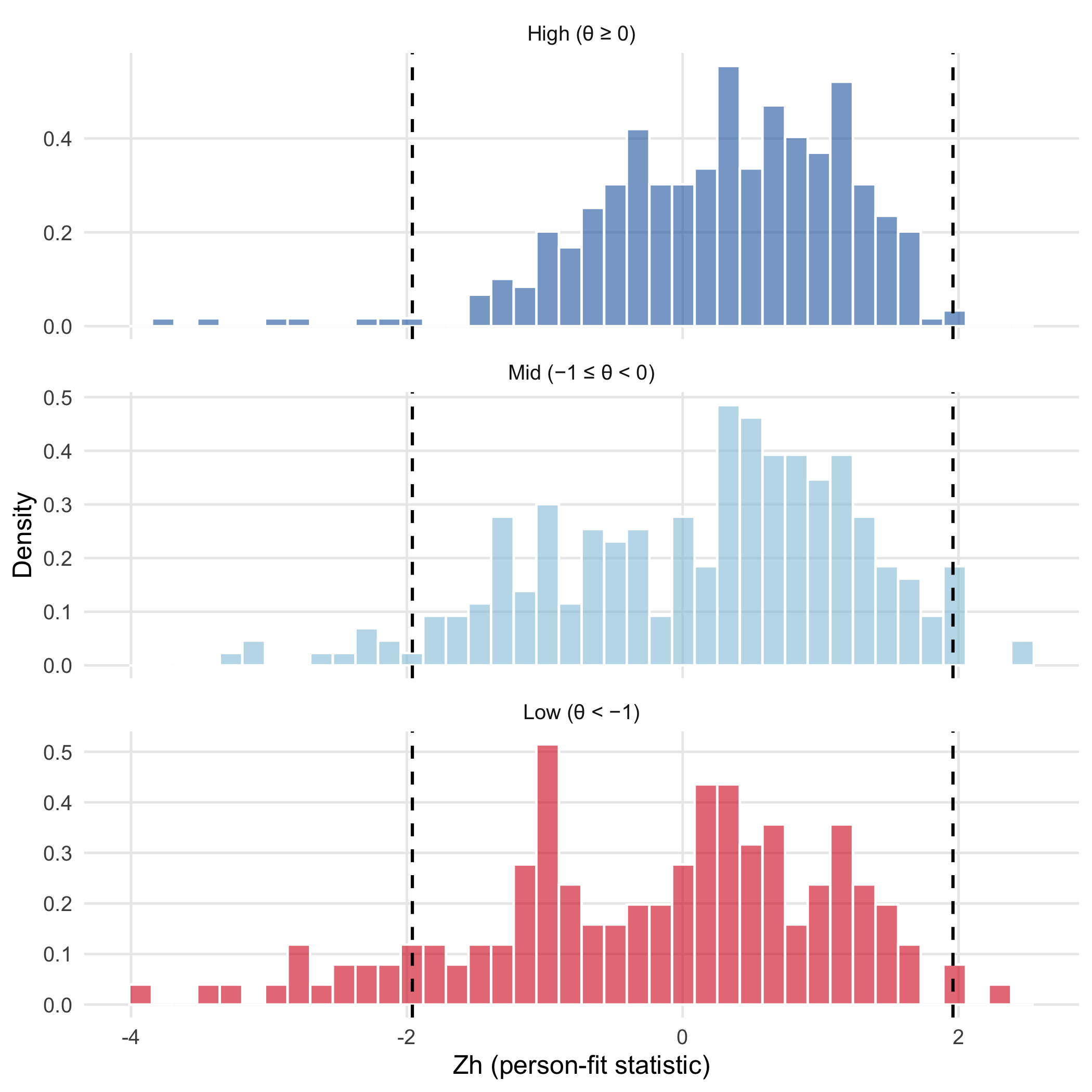}
    \caption{Zh distribution by IRT ability group. Low-ability pairs ($\theta < -1$, bottom panel) show substantially wider and more left-skewed distributions than high- or mid-ability pairs, indicating elevated rates of statistically anomalous response patterns.}
    \label{fig:zh_by_group}
\end{minipage}%
\hfill
%--- RIGHT minipage: Figure 2 ---
\begin{minipage}[b]{0.48\linewidth}
  \centering
  \includegraphics[width=\textwidth]{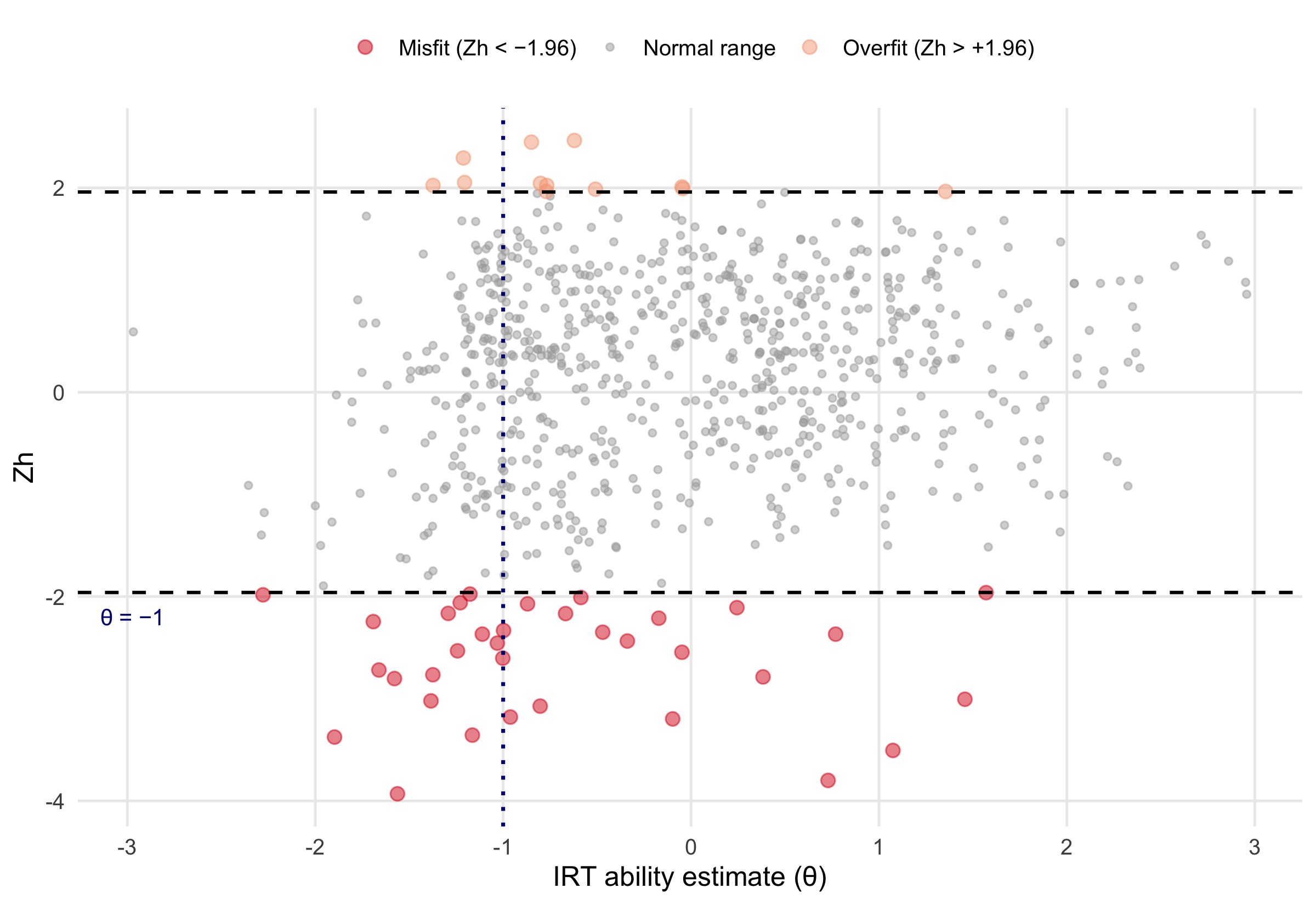}
  \caption{IRT ability ($\theta$) vs.\ person-fit statistic (Zh). Formally misfitting pairs (red, Zh $< -1.96$) appear across the full ability range, confirming that Zh carries diagnostic information orthogonal to $\theta$.}
  \label{fig:theta_zh_scatter}
\end{minipage}
\end{figure}

\begin{figure*}[!htbp]
    \centering
    \includegraphics[width=0.8\textwidth]{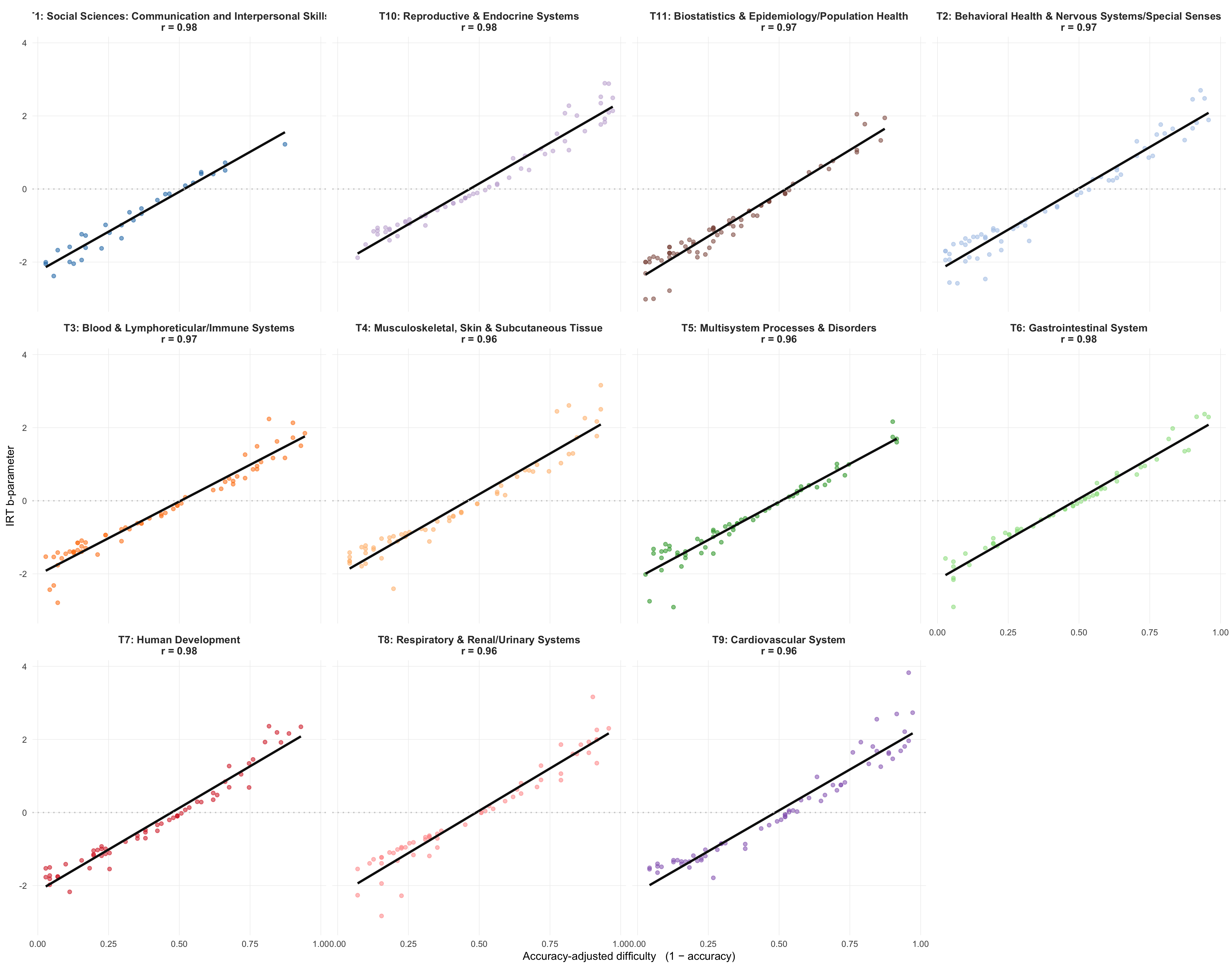}
    \caption{IRT $b$-parameter vs.\ accuracy-based difficulty across all 11 medical topics ($r = 0.96$--$0.98$ per topic). The tight and consistent linear relationship across every domain confirms that IRT difficulty is robustly grounded in item-level empirical hardness.}
    \label{fig:b_per_topic}
\end{figure*}

\paragraph{IRT item difficulty is grounded in item-level accuracy.}
Figure~\ref{fig:b_per_topic} shows the relationship between IRT $b$-parameters and accuracy-based difficulty (1 $-$ proportion correct) across all 11 USMLE topics. Pearson correlations range from $r = 0.96$ to $r = 0.98$ in every topic panel, confirming that IRT difficulty aligns tightly with the intuitive notion of item hardness. The added value of $b$ over simple proportion-correct lies in its joint estimation across all models and items simultaneously, yielding estimates that account for item discrimination and are more statistically stable for items answered by few or many models.

\begin{table*}[!hbpt]
\centering
\caption{\textbf{Top Cost- and Time-Normalized Performance (Ability).} Models are ranked by $\theta$/\$ (mean IRT ability per total cost). C and T refer to total cost (USD) and mean time (seconds), respectively. The other metric is Ability/s (mean $\theta$ per second), computed using mean response time (s/question).}
% Ranks are reported for $\theta$ and $\theta$/s among all 71 LLMs (larger is better).}
\label{tab:cost_time_performance_ability}
\small
\begin{tabular}{lrrrrrrr}
\toprule
\textbf{Model} & \textbf{$\theta \uparrow$} & \textbf{C (\$) $\downarrow$} & \textbf{T (s) $\downarrow$} & \textbf{$\theta/\$ \uparrow$} & \textbf{$\theta$/s $\uparrow$} & \textbf{Rank $\theta$} & \textbf{Rank $\theta$/s} \\
\midrule
meta-llama/llama-3.3-70b-instruct & 0.726 & 0.01 & 1.99 & 72.636 & 0.365 & 17 & 5 \\
meta-llama/llama-4-maverick       & 1.045 & 0.04 & 2.05 & 26.13  & 0.510 & 12  & 3 \\
moonshotai/kimi-k2                & 1.119 & 0.05 & 4.40 & 22.38  & 0.254 & 10 & 6 \\
openai/gpt-oss-120b               & 1.710 & 0.10 & 2.81 & 17.10  & 0.608 & 4  & 2 \\
google/gemini-2.5-flash           & 1.068 & 0.08 & 6.97 & 13.35  & 0.153 & 11 & 7 \\
openai/gpt-5-nano                 & 1.383 & 0.11 & 31.20& 12.57  & 0.044 & 6  & 9 \\
deepseek/deepseek-chat-v3.1       & 0.688 & 0.05 & 6.63 & 13.76  & 0.104 & 18 & 8 \\
openai/gpt-oss-20b                & 0.923 & 0.08 & 1.98 & 11.54  & 0.466 & 14 & 4 \\
openai/gpt-4.1-mini               & 0.985 & 0.11 & 25.87& 8.96   & 0.038 & 13 & 10 \\
qwen/qwen3-30b-a3b                & 0.871 & 0.10 & 1.06 & 8.71   & 0.822 & 15 & 1 \\
\bottomrule
\end{tabular}
\end{table*}

\begin{figure}[h]
    \centering
    \includegraphics[width=0.8\textwidth]{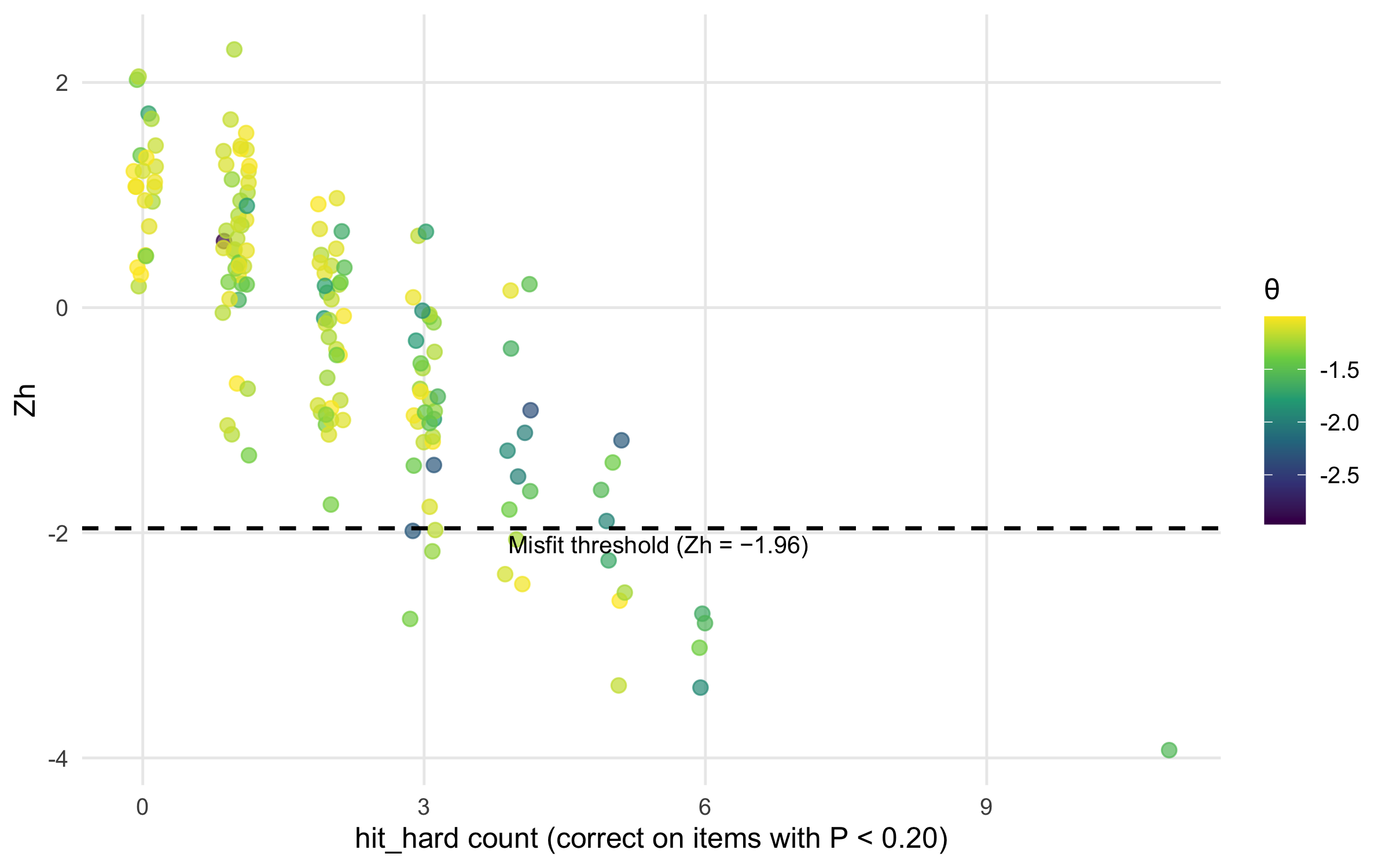}
    \caption{Anomalous hard-item hits (hit\_hard) vs.\ Zh among low-ability model--topic pairs ($\theta < -1$). Points below the dashed misfit threshold (Zh $= -1.96$) concentrate at higher hit\_hard counts, identifying format brittleness as the primary driver of response-pattern misfit in this group.}
    \label{fig:hithard_zh}
\end{figure}

\paragraph{Person-fit (Zh) identifies anomalous response patterns independently of ability.}
Figures~\ref{fig:zh_by_group}--\ref{fig:hithard_zh} characterize the person-fit statistic Zh and its diagnostic role. Figure~\ref{fig:zh_by_group} shows that high-ability model--topic pairs concentrate within the normal Zh range ($-1.96$ to $+1.96$), while low-ability pairs ($\theta < -1$) produce a substantially wider and more left-skewed distribution, indicating that weak models are disproportionately likely to exhibit statistically anomalous response patterns. Figure~\ref{fig:theta_zh_scatter} confirms that this is not a simple restatement of ability: formally misfitting pairs appear at mid and high $\theta$ values as well, making Zh an orthogonal diagnostic signal rather than a redundant proxy for accuracy. Figure~\ref{fig:hithard_zh} isolates the mechanism among low-ability pairs: model--topic pairs with high counts of correct responses on very hard items (hit\_hard $\geq 3$, items answered correctly by fewer than 20\% of models) fall predominantly below the Zh misfit threshold. This is the format brittleness signature — inconsistent successes on hard items that are statistically incompatible with the model's overall ability profile, and that accuracy-based evaluation would treat as genuine competence.
 
\section{Cost-Performance Trade-offs}
\label{sec:cost-performance}

Economic considerations play a crucial role in practical LLM deployment decisions. We analyze this by examining the relationship between model capability (mean IRT ability) and the associated financial cost and inference latency of the evaluation. Top cost- and time-normalized performance (based on mean IRT ability) is shown in Table~\ref{tab:cost_time_performance_ability}.
Practical deployment considerations reveal a clear Pareto frontier~\citep{jahan2016multi} in the cost-performance landscape. Considering the joint objectives of (i) mean IRT ability $(\theta)$, (ii) cost-normalized performance $(\theta/\$)$, and (iii) time-normalized performance $(\theta/\text{s})$, the non-dominated set comprises four models: \texttt{GPT-oss-120b}, \texttt{Llama-4-maverick}, \texttt{Llama-3.3-70b-instruct}, and \texttt{Qwen3-30b-a3b}. Each occupies a distinct operating point on the efficiency–performance landscape: \texttt{GPT-oss-120b} attains the highest absolute ability (strongest $\theta$); \texttt{Llama-3.3-70b-instruct} is the cost leader (maximizing $ \theta/\$ $); \texttt{Qwen3-30b-a3b} is the latency/throughput leader (maximizing $\theta/\text{s}$); and \texttt{Llama-4-maverick} provides a balanced trade-off—high $\theta$ with favorable $ \theta/\$ $ and $\theta/\text{s}$.

% \textbf{Operational metrics} capture deployment-relevant telemetry such as API costs and inference latency, supporting multi-objective optimization that balances capability, efficiency, and economic constraints. 

% \clearpage
% \onecolumn                 % longtable works best in one column

\section{Complete List of All Models}
\label{app:all_models}

\begin{table*}[!hpb]
\centering
\caption{Summary Statistics of Sampled Large Language Models (N=71)}
\label{tab:model_statistics}
\footnotesize
\begin{tabular}{llr|llr}
\toprule
\multicolumn{3}{c}{\textbf{Categorical Attributes}} &
\multicolumn{3}{c}{\textbf{Numerical Attributes}} \\
\cmidrule(r){1-3} \cmidrule(l){4-6}
\textbf{Attribute} & \textbf{Category} & \textbf{Count (\%)} &
\textbf{Metric} & \textbf{Statistic} & \textbf{Value} \\
\midrule
Access Type & Proprietary & 46 (64.8) & Context Length & Min & 2,824 \\
 & Open Source & 26 (36.6) &  & Q1 & 32,768 \\
 &  &  &  & Median & 128,000 \\
Model Size & Large & 18 (25.4) &  & Q3 & 172,880 \\
 & Medium & 19 (26.8) &  & Max & 2,000,000 \\
 & Small & 25 (35.2) &  &  &  \\
 & Unknown & 9 (12.7) & Completion Price & Min & 0.006 \\
 &  &  & (\$/1M tokens) & Q1 & 0.135 \\
Reasoning & Chat & 35 (49.3) &  & Median & 0.400 \\
 & Reasoning & 20 (28.2) &  & Q3 & 1.720 \\
 & Base & 16 (22.5) &  & Max & 75.000 \\
 &  &  &  &  &  \\
Modality & Text-only & 45 (63.4) & Prompt Price & Min & 0.003 \\
 & Multimodal & 26 (36.6) & (\$/1M tokens) & Q1 & 0.050 \\
 &  &  &  & Median & 0.150 \\
Top Vendors & Google, OpenAI & 9 each &  & Q3 & 0.712 \\
 & Meta, Mistral & 8 each &  & Max & 15.000 \\
 & Cohere, Qwen & 4 each &  &  &  \\
 & Anthropic, Deepseek, Microsoft & 3 each &  &  &  \\
 & Others (21 vendors) & 1--2 each &  &  &  \\
\bottomrule
\end{tabular}
\end{table*}

{\scriptsize
\begin{longtable}{p{3.5cm}|p{1.5cm}|p{1.3cm}|p{1.4cm}|p{1.3cm}|p{1cm}|p{1cm}}
% {p{3.8cm}|p{1.4cm}|p{1.1cm}|p{1.2cm}|p{1.2cm}|p{1cm}|p{1cm}}
% \caption{Evaluated Large Language Models with Mean IRT Ability (71 Models)}
\caption{Complete List of 71 Evaluated Large Language Models (ranked by mean IRT ability)}
\label{tab:71_llms_with_metadata}\\
\toprule
\textbf{Model Name} & \textbf{Vendor} & \textbf{Size} & \textbf{Access} &
\textbf{Release} & \textbf{Ability} & \textbf{Accuracy} \\
\midrule
\endfirsthead

\multicolumn{7}{c}{{\bfseries Table \thetable{} (continued)}}\\
\toprule
\textbf{Model Name} & \textbf{Vendor} & \textbf{Size} & \textbf{Access} &
\textbf{Release} & \textbf{Ability} & \textbf{Accuracy} \\
\midrule
\endhead

\midrule \multicolumn{7}{r}{{Continued on next page}} \\
\endfoot
\bottomrule
\endlastfoot

OpenAI GPT-5 & OpenAI & Unknown & Proprietary & Aug 2025 & 2.572 & 74.5\% \\
Google Gemini 2.5 Pro & Google & Large & Proprietary & Jun 2025 & 2.013 & 68.8\% \\
OpenAI Codex Mini & OpenAI & Small & Proprietary & May 2025 & 1.966 & 66.9\% \\
OpenAI GPT-OSS-120B & OpenAI & Large & Open & Aug 2025 & 1.738 & 63.5\% \\
Anthropic Claude Sonnet 4 & Anthropic & Large & Proprietary & May 2025 & 1.508 & 58.0\% \\
OpenAI GPT-5 Nano & OpenAI & Small & Proprietary & Aug 2025 & 1.383 & 61.2\% \\
OpenAI GPT-4o & OpenAI & Large & Proprietary & May 2024 & 1.264 & 58.5\% \\
xAI Grok 3 Mini & xAI & Small & Proprietary & Jun 2025 & 1.259 & 60.7\% \\
Anthropic Claude 3.7 Sonnet & Anthropic & Large & Proprietary & Feb 2025 & 1.117 & 57.6\% \\
Google Gemini 2.5 Flash & Google & Large & Proprietary & Jun 2025 & 1.086 & 57.4\% \\
MoonshotAI Kimi K2 & MoonshotAI & Large & Proprietary & Jul 2025 & 1.086 & 57.6\% \\
OpenAI GPT-4.1 Mini & OpenAI & Medium & Proprietary & Apr 2025 & 1.050 & 56.9\% \\
Qwen Qwen3 30B A3B & Qwen & Medium & Open & Apr 2025 & 0.916 & 56.4\% \\
OpenAI GPT-OSS-20B & OpenAI & Medium & Open & Aug 2025 & 0.914 & 56.7\% \\
Meta Llama 3.3 70B & Meta & Large & Open & Dec 2024 & 0.829 & 55.7\% \\
xAI Grok 2 Vision 1212 & xAI & Large & Proprietary & Dec 2024 & 0.756 & 54.1\% \\
DeepSeek DeepSeek V3.1 & DeepSeek & Large & Open & Aug 2025 & 0.688 & 53.8\% \\
Google Gemini Pro 1.5 & Google & Large & Proprietary & Apr 2024 & 0.597 & 53.1\% \\
Shisa AI Shisa V2 & Shisa AI & Large & Proprietary & Apr 2025 & 0.567 & 52.6\% \\
Anthracite Magnum V4 72B & Anthracite & Large & Proprietary & Oct 2024 & 0.555 & 51.4\% \\
Anthropic Claude 3 Opus & Anthropic & Large & Proprietary & Mar 2024 & 0.512 & 50.6\% \\
Meta Llama 4 Scout & Meta & Medium & Proprietary & Apr 2025 & 0.507 & 52.6\% \\
Nous Hermes 3 40B & NousResearch & Large & Proprietary & Aug 2024 & 0.491 & 51.0\% \\
Qwen Qwen Plus & Qwen & Small & Proprietary & Feb 2025 & 0.434 & 51.3\% \\
MiniMax MiniMax-01 & MiniMax & Large & Proprietary & Jan 2025 & 0.421 & 50.5\% \\
Z.AI GLM 4 32B & Z.AI & Medium & Proprietary & Jul 2025 & 0.419 & 51.0\% \\
THUDM GLM 4 32B & THUDM & Medium & Open & Apr 2025 & 0.413 & 50.7\% \\
Arcee AI Virtuoso Large & Arcee AI & Large & Proprietary & May 2025 & 0.347 & 49.6\% \\
Google Gemini 2.5 Flash Lite & Google & Small & Proprietary & Jun 2025 & 0.305 & 48.0\% \\
TheDrummer Skyfall 36B V2 & TheDrummer & Medium & Proprietary & Mar 2025 & 0.246 & 48.5\% \\
Nous DeepHermes 3 Mistral 24B & NousResearch & Medium & Proprietary & May 2025 & 0.245 & 47.7\% \\
Mistral Small 3.2 24B & Mistral AI & Medium & Proprietary & Jun 2025 & 0.239 & 46.7\% \\
Microsoft Phi 4 Reasoning Plus & Microsoft & Medium & Proprietary & May 2025 & 0.112 & 46.0\% \\
Qwen Qwen2.5 VL 32B & Qwen & Medium & Open & Mar 2025 & -0.055 & 44.9\% \\
Google Gemma 3 27B & Google & Medium & Open & Mar 2025 & -0.065 & 44.9\% \\
OpenAI GPT-4.1 Nano & OpenAI & Small & Proprietary & Apr 2025 & -0.071 & 44.7\% \\
Inflection 3 Productivity & Inflection & Unknown & Proprietary & Oct 2024 & -0.072 & 42.6\% \\
DeepSeek R1 Distill Qwen 14B & DeepSeek & Medium & Open & Jan 2025 & -0.116 & 45.8\% \\
DeepSeek R1 Distill Qwen3 8B & DeepSeek & Small & Open & May 2025 & -0.116 & 47.9\% \\
Raifle SorcererLM 8x22B & Raifle & Large & Proprietary & Nov 2024 & -0.195 & 41.9\% \\
Amazon Nova Micro V1 & Amazon & Small & Proprietary & Dec 2024 & -0.331 & 40.6\% \\
Microsoft Phi 4 Multimodal & Microsoft & Small & Proprietary & Mar 2025 & -0.385 & 40.7\% \\
Meta Llama 3 8B & Meta & Small & Open & Apr 2024 & -0.388 & 42.3\% \\
Sao10K Llama 3 8B Lunaris & Sao10K & Small & Open & Aug 2024 & -0.395 & 42.1\% \\
Meta Llama 3.2 11B Vision & Meta & Medium & Open & Sep 2024 & -0.409 & 41.1\% \\
Mistral Pixtral 12B & Mistral AI & Medium & Open & Sep 2024 & -0.422 & 37.8\% \\
Bytedance UI-TARS 1.5 7B & Bytedance & Small & Proprietary & Jul 2025 & -0.498 & 38.5\% \\
Qwen Qwen VL Plus & Qwen & Medium & Open & Feb 2025 & -0.507 & 38.5\% \\
Inception Mercury Coder & Inception & Unknown & Proprietary & Apr 2025 & -0.521 & 36.2\% \\
Microsoft Phi 3 Mini 128K & Microsoft & Small & Open & May 2024 & -0.555 & 37.7\% \\
Liquid LFM 7B & Liquid & Small & Proprietary & Jan 2025 & -0.587 & 36.1\% \\
Cohere Command R 08-2024 & Cohere & Unknown & Proprietary & Aug 2024 & -0.611 & 35.9\% \\
Morph Morph V3 Large & Morph & Large & Proprietary & Jul 2025 & -0.634 & 35.5\% \\
Meta Llama 3.2 3B & Meta & Small & Open & Sep 2024 & -0.658 & 36.2\% \\
Baidu ERNIE 4.5 21B A3B & Baidu & Medium & Proprietary & Aug 2025 & -0.661 & 39.7\% \\
Mistral Ministral 8B & Mistral AI & Small & Open & Oct 2024 & -0.682 & 34.6\% \\
Google Gemini Flash 1.5 8B & Google & Small & Proprietary & Oct 2024 & -0.689 & 36.1\% \\
Mistral Codestral 2508 & Mistral AI & Medium & Proprietary & Aug 2025 & -0.698 & 33.0\% \\
Mistral Codestral 2501 & Mistral AI & Small & Open & Jan 2025 & -0.717 & 32.8\% \\
AI21 Jamba Mini 1.7 & AI21 & Small & Proprietary & Aug 2025 & -0.720 & 35.0\% \\
Cohere Command R7B 12-2024 & Cohere & Unknown & Proprietary & Dec 2024 & -0.736 & 34.7\% \\
Cohere Command R & Cohere & Medium & Proprietary & Mar 2024 & -0.749 & 32.4\% \\
Alpindale Goliath 120B & Alpindale & Large & Open & Nov 2023 & -0.751 & 31.8\% \\
Meta LlamaGuard 2 8B & Meta & Small & Open & May 2024 & -0.758 & 32.9\% \\
OpenAI GPT-3.5 Turbo Instruct & OpenAI & Small & Proprietary & Sep 2023 & -0.806 & 32.5\% \\
Google Gemma 3 4B & Google & Small & Open & Mar 2025 & -0.867 & 32.7\% \\
Mistral Tiny & Mistral AI & Small & Open & Jan 2024 & -1.166 & 32.6\% \\
Cohere Command & Cohere & Unknown & Proprietary & Mar 2024 & -1.295 & 29.2\% \\
Gryphe MythoMax L2 13B & Gryphe & Medium & Open & Jul 2023 & -1.410 & 27.7\% \\
Mistral 7B Instruct v0.1 & Mistral AI & Small & Open & Sep 2023 & -1.586 & 25.0\% \\
AlfredPros CodeLlama 7B & AlfredPros & Small & Proprietary & Apr 2025 & -1.943 & 14.8\% \\
\end{longtable}
}

%%%%%% This section serves as the Inversion summary table. 
\begingroup\fontsize{9}{11}\selectfont

\begin{longtable}[t]{>{\raggedleft\arraybackslash}p{0.6cm}>{\raggedright\arraybackslash}p{6.5cm}>{\raggedright\arraybackslash}p{5.5cm}}
\caption{\label{tab:dir_inversion}Difficulty-insensitive responding (DIR) models: all inversion pairs and magnitude of change.}\\
\toprule
Rank & Model & Inversion Pair\\
\midrule
\endfirsthead
\caption[]{Difficulty-insensitive responding (DIR) models: all inversion pairs and magnitude of change. \textit{(continued)}}\\
\toprule
Rank & Model & Inversion Pair\\
\midrule
\endhead

\endfoot
\bottomrule
\endlastfoot
3 & openai/codex-mini & Ext.\ Easy $\rightarrow$ Easy (+9.6\%)\\
\cellcolor{gray!10}{4} & \cellcolor{gray!10}{openai/gpt-oss-120b} & \cellcolor{gray!10}{Ext.\ Easy $\rightarrow$ Easy (+2.4\%)}\\
5 & anthropic/claude-sonnet-4 & Ext.\ Easy $\rightarrow$ Easy (+1.9\%)\\
\cellcolor{gray!10}{7} & \cellcolor{gray!10}{openai/gpt-4o} & \cellcolor{gray!10}{Ext.\ Easy $\rightarrow$ Easy (+3.8\%)}\\
11 & google/gemini-2.5-flash & Ext.\ Easy $\rightarrow$ Easy (+2.4\%)\\
\addlinespace
\cellcolor{gray!10}{12} & \cellcolor{gray!10}{meta-llama/llama-4-maverick} & \cellcolor{gray!10}{Ext.\ Easy $\rightarrow$ Easy (+6.3\%)}\\
13 & openai/gpt-4.1-mini & Ext.\ Easy $\rightarrow$ Easy (+7.2\%)\\
\cellcolor{gray!10}{14} & \cellcolor{gray!10}{openai/gpt-oss-20b} & \cellcolor{gray!10}{Ext.\ Easy $\rightarrow$ Easy (+0.0\%)}\\
17 & meta-llama/llama-3.3-70b-instruct & Ext.\ Easy $\rightarrow$ Easy (+0.5\%)\\
\cellcolor{gray!10}{18} & \cellcolor{gray!10}{deepseek/deepseek-chat-v3.1} & \cellcolor{gray!10}{Ext.\ Easy $\rightarrow$ Easy (+3.4\%)}\\
\addlinespace
19 & google/gemini-2.5-flash-lite-preview-06-17 & Ext.\ Easy $\rightarrow$ Easy (+2.4\%)\\
 &  & Hard $\rightarrow$ Ext.\ Hard (+0.2\%)\\
\cellcolor{gray!10}{20} & \cellcolor{gray!10}{google/gemini-pro-1.5} & \cellcolor{gray!10}{Hard $\rightarrow$ Ext.\ Hard (+2.8\%)}\\
22 & shisa-ai/shisa-v2-llama3.3-70b & Ext.\ Easy $\rightarrow$ Easy (+4.3\%)\\
\cellcolor{gray!10}{23} & \cellcolor{gray!10}{meta-llama/llama-4-scout} & \cellcolor{gray!10}{Ext.\ Easy $\rightarrow$ Easy (+1.0\%)}\\
\addlinespace
24 & anthropic/claude-3-opus & Ext.\ Easy $\rightarrow$ Easy (+0.5\%)\\
\cellcolor{gray!10}{25} & \cellcolor{gray!10}{z-ai/glm-4-32b} & \cellcolor{gray!10}{Ext.\ Easy $\rightarrow$ Easy (+1.0\%)}\\
26 & thudm/glm-4-32b & Ext.\ Easy $\rightarrow$ Easy (+1.0\%)\\
\cellcolor{gray!10}{30} & \cellcolor{gray!10}{google/gemini-2.5-flash-lite} & \cellcolor{gray!10}{Ext.\ Easy $\rightarrow$ Easy (+8.2\%)}\\
31 & nousresearch/deephermes-3-mistral-24b-preview & Hard $\rightarrow$ Ext.\ Hard (+4.0\%)\\
\addlinespace
\cellcolor{gray!10}{32} & \cellcolor{gray!10}{mistralai/mistral-small-3.2-24b-instruct} & \cellcolor{gray!10}{Ext.\ Easy $\rightarrow$ Easy (+0.0\%)}\\
34 & deepseek/deepseek-r1-distill-qwen-14b & Hard $\rightarrow$ Ext.\ Hard (+2.8\%)\\
\cellcolor{gray!10}{35} & \cellcolor{gray!10}{microsoft/phi-4-reasoning-plus} & \cellcolor{gray!10}{Ext.\ Easy $\rightarrow$ Easy (+7.7\%)}\\
36 & openai/gpt-4.1-nano & Ext.\ Easy $\rightarrow$ Easy (+3.4\%)\\
\cellcolor{gray!10}{39} & \cellcolor{gray!10}{inflection/inflection-3-productivity} & \cellcolor{gray!10}{Ext.\ Easy $\rightarrow$ Easy (+0.1\%)}\\
\addlinespace
41 & amazon/nova-micro-v1 & Hard $\rightarrow$ Ext.\ Hard (+1.4\%)\\
\cellcolor{gray!10}{46} & \cellcolor{gray!10}{mistralai/pixtral-12b} & \cellcolor{gray!10}{Hard $\rightarrow$ Ext.\ Hard (+0.0\%)}\\
47 & bytedance/ui-tars-1.5-7b & Hard $\rightarrow$ Ext.\ Hard (+2.7\%)\\
\cellcolor{gray!10}{49} & \cellcolor{gray!10}{qwen/qwen-vl-plus} & \cellcolor{gray!10}{Hard $\rightarrow$ Ext.\ Hard (+0.0\%)}\\
50 & inception/mercury-coder & Hard $\rightarrow$ Ext.\ Hard (+1.3\%)\\
\addlinespace
\cellcolor{gray!10}{53} & \cellcolor{gray!10}{meta-llama/llama-3.2-3b-instruct} & \cellcolor{gray!10}{Hard $\rightarrow$ Ext.\ Hard (+19.8\%)}\\
58 & mistralai/codestral-2508 & Hard $\rightarrow$ Ext.\ Hard (+0.0\%)\\
\cellcolor{gray!10}{60} & \cellcolor{gray!10}{openai/gpt-3.5-turbo-instruct} & \cellcolor{gray!10}{Hard $\rightarrow$ Ext.\ Hard (+2.7\%)}\\
65 & cohere/command-r & Hard $\rightarrow$ Ext.\ Hard (+1.3\%)\\
\cellcolor{gray!10}{69} & \cellcolor{gray!10}{gryphe/mythomax-l2-13b} & \cellcolor{gray!10}{Hard $\rightarrow$ Ext.\ Hard (+2.6\%)}\\
\addlinespace
71 & alfredpros/codellama-7b-instruct-solidity & Easy $\rightarrow$ Medium (+4.1\%)\\*
\end{longtable}
\endgroup{}

\end{document}